\documentclass{article}

\usepackage{arxiv}

\usepackage[utf8]{inputenc}
\usepackage[T1]{fontenc}
\IfFileExists{lmodern.sty}{\usepackage{lmodern}}{}
\usepackage{amsmath}
\usepackage{amssymb}
\usepackage{graphicx}
\usepackage{booktabs}
\usepackage{array}
\usepackage{longtable}
\usepackage{calc}
\providecommand{\real}[1]{#1}

\usepackage{pdflscape}
\usepackage{xcolor}
\usepackage{microtype}
\usepackage[numbers,sort&compress]{natbib}
\usepackage{hyperref}
\usepackage{flafter}
\hypersetup{colorlinks=true, linkcolor=blue, citecolor=blue, urlcolor=blue}

\renewcommand{\shorttitle}{Portable EHR Representations}
\renewcommand{\undertitle}{}
\renewcommand{\headeright}{Preprint}

\title{PORTER: Language-Grounded Event Representations for Portable Structured EHR Foundation Models}

\author{%
  Lin Lawrence Guo\\
  Child Health Evaluative Sciences\\
  The Hospital for Sick Children\\
  Toronto, Canada
  \And
  Adam Paul Yan\\
  Child Health Evaluative Sciences\\
  Division of Haematology/Oncology\\
  The Hospital for Sick Children\\
  Toronto, Canada
  \AND
  Emily Vettese\\
  Child Health Evaluative Sciences\\
  The Hospital for Sick Children\\
  Toronto, Canada
  \And
  Lillian Sung\thanks{Corresponding author: \texttt{lillian.sung@sickkids.ca}}\\
  Child Health Evaluative Sciences\\
  Division of Haematology/Oncology\\
  The Hospital for Sick Children\\
  Toronto, Canada
}

\date{}

\begin{document}
\maketitle

\begin{abstract}
Most electronic health record (EHR) foundation models encode clinical events as discrete event tokens from a fixed vocabulary and therefore cannot directly represent events containing unseen concepts or new combinations of concepts and attributes such as numeric values. This limits transfer across institutions and even across deployment pipelines within the same institution. We introduce PORTER (Portable EHR Representations), a language-grounded structured EHR foundation model that decouples event representation from this fixed vocabulary. PORTER represents events through their descriptions using a frozen text encoder, integrates numeric values through a dedicated pathway, and learns clinical dynamics over patient timelines with an autoregressively pretrained temporal backbone. Across 74 clinical prediction tasks at a pediatric hospital, PORTER matched the mean area-under-the-receiver-operating-characteristic curve (AUROC) of a fixed-vocabulary model with the same temporal backbone and pretraining objective. When the same patient timelines were rendered using event descriptions not seen during pretraining, PORTER transferred without retraining or vocabulary mapping, recovering 97.1\% of the mean AUROC of a model trained directly on the target vocabulary. When transferred to MIMIC, PORTER outperformed the fixed-vocabulary model, which dropped 69\% of events because their tokens were unseen. Mechanistic analyses showed cross-vocabulary transfer tracked preservation of patient-level representation geometry rather than the scale of the text encoder, and the numeric pathway improved sensitivity to magnitude without disrupting clinical concept identity. PORTER also achieved higher AUROC than a task-specific text serialization comparator, at 329-fold lower amortized compute. PORTER is a step toward vocabulary-independent EHR foundation models that reduce the need for vocabulary harmonization while preserving in-domain performance and enabling efficient cross-task reuse.
\end{abstract}

\section{INTRODUCTION}

Structured electronic health record (EHR) foundation models learn reusable patient representations through self-supervised pretraining on longitudinal sequences of clinical events.\cite{RN1,RN2,RN3,RN4,RN5,RN6,RN7,RN8} Each event is a timestamped occurrence of a clinical concept, such as a diagnosis or laboratory result, and may include structured attributes such as numeric values. These models have demonstrated robustness under temporal dataset shift\cite{RN9} and transfer across populations\cite{RN10} and institutions\cite{RN11}, although such transfer generally assumes that evaluation data can be encoded using event-token representations learned during pretraining.

In practice, the codes and descriptions used to represent clinical events vary across institutions, coding standards, and data pipelines. Most structured EHR foundation models convert each clinical event into a discrete event token from a fixed vocabulary learned during pretraining.\cite{RN12} These tokens may represent a clinical concept alone or a composite of the clinical concept and structured attributes, such as a discretized numeric bin.\cite{RN13} When deployment data include new clinical concepts, different local codes or names for the same concepts, or new concept-attribute combinations, those events produce tokens for which the model has no learned representation. Even within a single institution, models may be developed using retrospective datasets mapped to common data models such as the Observational Medical Outcomes Partnership Common Data Model (OMOP CDM), then deployed using production feeds with different local codes or naming conventions. Therefore, vocabulary shift can arise even when the patient population and clinical setting remain the same.

A portable structured EHR foundation model must preserve the meaning of a clinical event even when that event is described using different codes or descriptions. Standard vocabularies and common data models reduce this heterogeneity, but mapping to them is labor-intensive, error-prone, and does not eliminate differences in concept coverage or local implementation. Numeric values introduce a second representation challenge. For measurement events, the clinical concept identifies what was measured, while the numeric value and reference range determine magnitude and abnormality. A portable model must therefore represent both concept semantics and numeric values without tying either to a fixed token vocabulary.

Text-based representations have emerged as a way to reduce the dependence of structured EHR models on fixed vocabularies and manual harmonization. One line of work serializes patient timelines into natural language and uses a large language model or text encoder to produce patient-level representations.\cite{RN14,RN15,RN16} This approach avoids a fixed event-token vocabulary, but requires the model to infer concept semantics, temporal relationships, and numeric meaning from the serialized text.\cite{RN17} When serialization is task-specific, patient representations must also be recomputed for each downstream task. A second line of work operates at the code or event level by representing clinical concepts through their descriptions or by learning aligned code representations across sites.\cite{RN18,RN19,RN20,RN21,RN22,RN23} These methods have shown that text-based code or event representations can improve portability across vocabularies, schemas, institutions, and languages. They differ in whether the text encoder is frozen or updated during training, which determines how much an event description\textquotesingle s representation is shaped by the training vocabulary. A frozen pretrained encoder applies the same semantic mapping to descriptions from any vocabulary, including vocabularies not seen during training, without modifying model weights\cite{RN18}. Some also use self-supervised objectives, but as initialization for supervised fine-tuning rather than to learn task-agnostic patient representations reused across downstream tasks.\cite{RN22} Across existing frameworks, numeric values are commonly discarded, rendered as text, discretized, or represented through token-level schemes. Existing approaches therefore have not fully combined vocabulary-independent inputs, autoregressive pretraining for reusable task-agnostic patient representations, and explicit representation of numeric values within a single structured EHR foundation model (Supplementary Table S1).

We introduce PORTER (Portable EHR Representations), a language-grounded structured EHR foundation model that separates concept semantics, numeric values, and temporal dynamics into distinct components. Rather than serializing full patient histories into text, PORTER applies language grounding at the clinical event level. Each event is paired with a natural-language description of its underlying clinical concept, which is processed by a frozen text encoder to provide vocabulary-independent concept representations. Because this text encoder is applied once per unique event description rather than repeatedly across patient histories, these representations can be cached and reused across patients and downstream tasks. For events containing numeric values, a separate learned pathway encodes numeric magnitude and relative abnormality directly, rather than rendering numeric values as text. Feature-wise linear modulation (FiLM)\cite{RN30} then uses this numeric information to modulate the text-derived concept representation and produce the event input representation. Finally, a temporal backbone learns clinical dynamics from sequences of these event input representations through autoregressive pretraining. After pretraining, the backbone is frozen and produces reusable patient representations that downstream tasks use through linear probes. At inference, PORTER can represent clinical concepts from vocabularies not seen during pretraining without retraining or explicit vocabulary mapping, provided they can be rendered as interpretable event descriptions.

We evaluate PORTER across three settings using a fixed-vocabulary EHR foundation model (Fixed-Vocab FM) matched on backbone architecture, pretraining objective, and training schedule as the primary comparator. First, we test whether PORTER matches Fixed-Vocab FM in-domain across 74 clinical prediction tasks. Second, we evaluate cross-vocabulary transfer within the same institution by applying pretrained PORTER to patient timelines where event descriptions are derived from institutional EHR terminology rather than the OMOP-derived descriptions used during pretraining. This setting isolates vocabulary shift while holding patient timelines fixed and cannot be directly supported by Fixed-Vocab FM without vocabulary mapping or retraining. Third, we evaluate cross-site transfer to MIMIC, where vocabulary, population, and clinical setting differ from pretraining. We additionally ablate the numeric pathway and text encoder choice across evaluation settings and compare PORTER with a patient-level text serialization comparator. This study makes the following contributions:

\begin{itemize}
\item
  We introduce PORTER, a language-grounded structured EHR foundation model that uses event-description inputs instead of fixed-vocabulary input embeddings. PORTER pairs a frozen pretrained text encoder, which yields vocabulary-independent concept representations that are cached and reused, with a dedicated numeric pathway that integrates magnitude through FiLM, and learns clinical dynamics with an autoregressive temporal backbone that is frozen after pretraining and reused across tasks through linear probes.
\item
  We evaluate PORTER using a controlled cross-vocabulary design that holds patients and downstream task labels fixed while changing only the event-description naming system. PORTER transferred to unseen event descriptions without retraining or vocabulary mapping, recovering 97.1\% of the AUROC of a target-vocabulary reference model. Text encoder ablations showed that cross-vocabulary performance tracked preservation of patient-level representation geometry rather than encoder scale.
\item
  PORTER matched an architecture-matched Fixed-Vocab FM across 74 in-domain tasks and improved transfer on 31 of 36 tasks at an external site, where Fixed-Vocab FM dropped 69\% of events because their tokens were unseen.
\item
  PORTER's dedicated numeric pathway improved sensitivity to numeric magnitude compared with rendering values as text, while preserving clinical concept identity.
\item
  Compared with a task-specific patient-level text serialization comparator, PORTER achieved higher AUROC on 69 of 74 tasks, with lower amortized compute as reuse of task-agnostic patient representations increased.
\end{itemize}

\section{METHODS}

\subsection{Hospital Datasets}

This study used EHR data from The Hospital for Sick Children (SickKids), a tertiary pediatric hospital, as the primary development site, and MIMIC, derived from Beth Israel Deaconess Medical Center (BIDMC), an adult academic medical center, for external evaluation. SickKids uses Epic (Epic Systems, Verona, WI) as its enterprise EHR.

The SickKids dataset was sourced from the SickKids Enterprise-wide Data in Azure Repository (SEDAR)\cite{RN24}, which consolidates EHR data from SickKids\textquotesingle{} Epic Clarity database into a clinically oriented, validated and standardized schema. EHR data were mapped to the Medical Event Data Standard (MEDS)\cite{RN25,RN26} format with clinical concepts standardized to Observational Medical Outcomes Partnership Common Data Model (OMOP CDM) ontologies. The MIMIC dataset (MIMIC-IV, version 1.0)\cite{RN27} contains de-identified EHR data from patients admitted to the intensive care unit or emergency department at BIDMC between 2008 and 2019. MIMIC data were mapped to the OMOP CDM using code provided by the Observational Health Data Sciences and Informatics MIMIC project\cite{RN28} and subsequently converted to MEDS format. As part of MIMIC\textquotesingle s de-identification process, patient timelines are shifted to an anchor year within a three-year window. To support consistent temporal splitting across SickKids and MIMIC, we deterministically assigned each patient a representative calendar year within their anchor group via hashing of the patient identifier.

Use of SEDAR data for this study was approved by the Research Ethics Board (REB) at SickKids (REB number: 1000074527). Use of MIMIC was approved under the oversight of the Institutional Review Board at BIDMC and is made publicly available on PhysioNet.\cite{RN29}

\subsection{Cohort Definition and Splitting}

The cohort selection process is summarized in Supplementary Figure S1. Pretraining cohorts were defined at the patient level. For SickKids, we included all patients in SEDAR over the study period, with clinical events spanning June 2, 2018 (EHR go-live) through April 7, 2026. Patients were excluded if they had missing date of birth. For MIMIC, all patients in the dataset were included. Within each dataset, patients were deterministically assigned to training (\textasciitilde90\%) and validation (\textasciitilde10\%) subsets via hashing of the patient identifier. For SickKids, pretraining used events through December 31, 2024 for training patients and through March 31, 2025 for validation patients. For MIMIC, pretraining used events through December 31, 2016 for training patients and through December 31, 2017 for validation patients.

Downstream evaluation cohorts were defined at the admission level. For SickKids, we included inpatient admissions where age at the prediction time was 28 days or older. For MIMIC, we included inpatient admissions where age at prediction time was 18 years or older. Admissions were assigned to training, validation, and test sets according to the prediction time, using the same calendar cutoffs applied to the corresponding pretraining cohorts. Patients with multiple admissions could contribute admissions to different temporal periods, reflecting a deployment setting in which previously observed patients may return for future admissions.

\subsection{Clinical Prediction Tasks}

The evaluation tasks were adapted from our previous study.\cite{RN13} At SickKids, we evaluated 74 clinical prediction tasks spanning six task families: transfusions, procedures, imaging, laboratory abnormalities, medication administrations, and clinical outcomes, with laboratory abnormalities defined using site-specific reference ranges. For MIMIC, we evaluated 36 tasks comprising an adapted subset of the SickKids clinical outcome and laboratory abnormality tasks, with laboratory abnormalities defined using MIMIC-specific reference ranges.

Prediction time was set to the end of the admission day at 23:59 for all tasks except 30-day readmission, for which it was set at 23:59 on the day before discharge. The prediction window extended until discharge for all tasks except long length of stay and readmission, which used fixed windows of 7 days after admission and 30 days after discharge, respectively. For each task, admissions in which death, discharge, or outcome occurred on or before the prediction time were excluded. The full task list and cohort sizes are provided in Supplementary Table S2.

\subsection{PORTER Architecture}

We define a clinical event as a timestamped EHR occurrence associated with a clinical concept, a natural-language event description, and, where applicable, numeric metadata such as value and reference range. Each event contributes one input representation to the temporal model. In contrast to fixed-vocabulary models that represent events by indexing a learned embedding table over discrete event tokens, PORTER generates event representations directly from these event descriptions and numeric metadata, making the input pathway vocabulary-independent.

PORTER comprises three components (Figure 1). First, a frozen text encoder converts each event description to a vocabulary-independent concept representation, which is cached and projected to the transformer hidden dimension. Second, for events with numeric measurements, a numeric pathway encodes measurement magnitude and relative abnormality, and feature-wise linear modulation (FiLM) integrates this numeric information by scaling and shifting the projected text-derived representation. Third, a temporal transformer backbone processes the fused event representations and learns longitudinal clinical dynamics through autoregressive next-event prediction. The following sections describe these components and the pretraining objective in detail.

\begin{figure}[t]
\centering
\includegraphics[width=\linewidth]{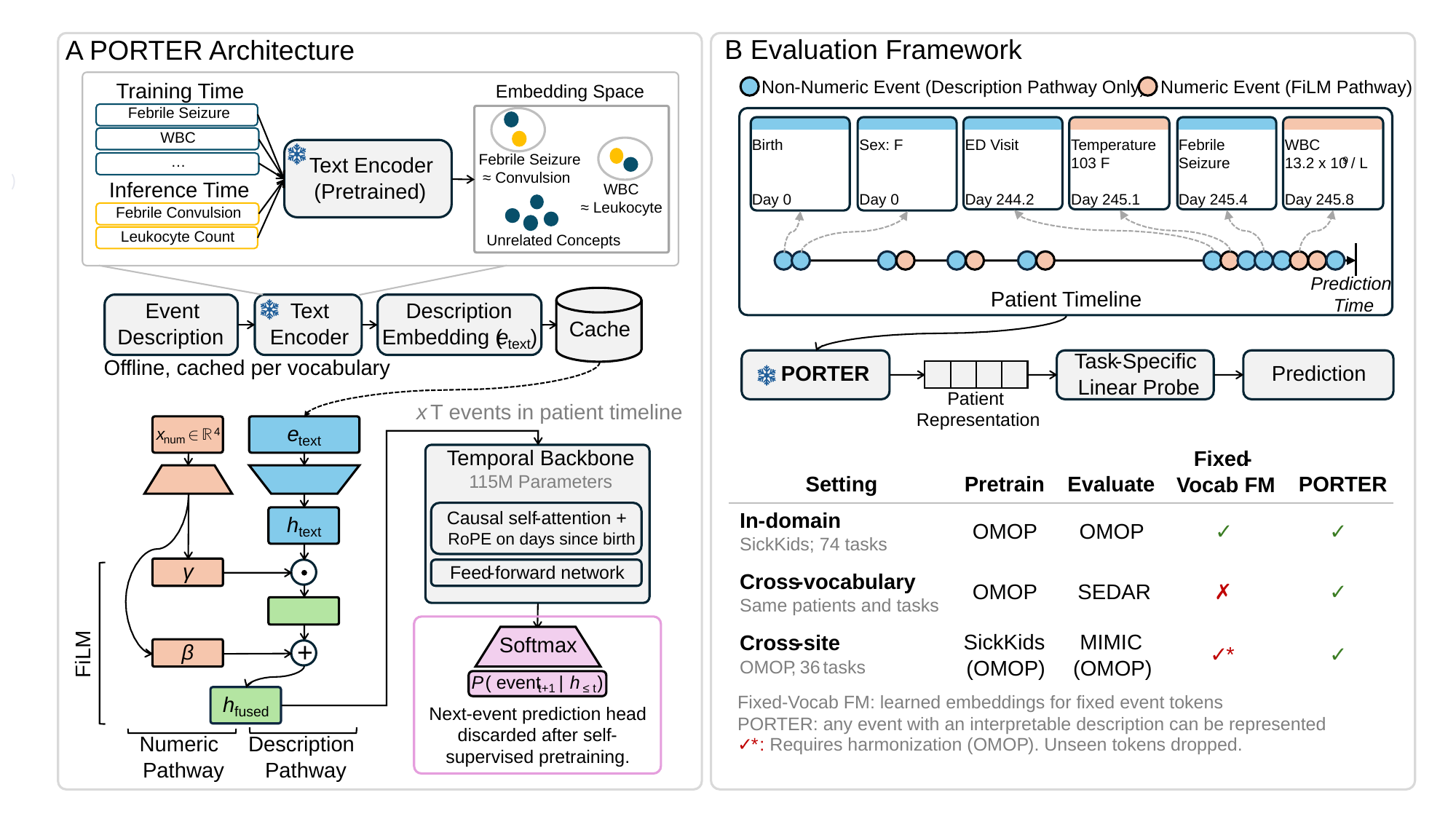}
\caption{PORTER architecture and evaluation framework. (A) PORTER represents each clinical event through an event-description pathway and, when applicable, a numeric pathway. A frozen pretrained text encoder, BioLORD, maps event descriptions to dense embeddings, $e_{\text{text}}$, which are projected into the backbone hidden dimension, $h_{\text{text}}$. For events with numeric values, a parallel numeric pathway encodes normalized numeric features through a learned multilayer perceptron and modulates the text-derived concept representation using feature-wise linear modulation, producing a fused event input representation, $h_{\text{fused}}$. Numeric values are normalized using institutional reference ranges where available, with log-transformed raw values as a fallback. Three binary indicators specify whether a numeric value is present and which normalization was applied. Event-description embeddings are computed once per unique event description and cached offline. A causal transformer backbone with 115M parameters and rotary position embeddings applied to patient age in days is pretrained using self-supervised next-event prediction over patient event sequences. The prediction head is discarded after pretraining. (B) Downstream evaluation extracts frozen patient representations at a task-specific prediction time and trains a task-specific linear probe. Three evaluation settings test increasing portability demands. In-domain evaluation uses the same institution and OMOP-derived event descriptions. Cross-vocabulary evaluation uses the same patients and labels but SEDAR-derived event descriptions. Cross-site evaluation transfers from SickKids to MIMIC. A conventional fixed-vocabulary foundation model, Fixed-Vocab FM, serves as the primary comparator. Fixed-Vocab FM can operate in-domain and can be evaluated cross-site when data are harmonized to OMOP, although unseen event tokens are dropped. It cannot directly support cross-vocabulary evaluation without vocabulary mapping or retraining. Abbreviations: PORTER, Portable EHR Representations; MLP, multilayer perceptron; OMOP, Observational Medical Outcomes Partnership; SEDAR, SickKids Enterprise-wide Data in Azure Repository; MIMIC, Medical Information Mart for Intensive Care; FiLM, feature-wise linear modulation; EHR, electronic health records; RoPE, rotary position embedding.}
\label{fig:architecture}
\end{figure}

\subsubsection{Language-Grounded Event Representation}

Each clinical event was represented by a templated event description, with numeric metadata retained for separate numeric encoding. Templates spanned nine categories: demographics, measurement, observation, condition, procedure, drug, note, specimen, and visit. Where relevant, templates incorporated additional structured EHR fields, including non-numeric measurement result categories and drug route. For example, measurement events were represented as "Measurement: \{concept name\}" or "Measurement: \{concept name\}. Result: \{non-numeric result concept name\}." For the primary PORTER setup, both the primary concept and attribute slots were populated using OMOP concept names. For cross-vocabulary evaluation, we constructed a separate set of event descriptions from analogous SEDAR concept and attribute names, allowing the same patient histories to be represented using institutional rather than OMOP terminology. Full templates and source columns are provided in Supplementary Table S3. Each event description was encoded once using a frozen text encoder and cached as a static event-description embedding table. We evaluated BioLORD-2023\cite{RN31}, BGE-M3\cite{RN32}, and Qwen3-Embedding-8B\cite{RN33} as candidate encoders and selected BioLORD-2023 for the primary PORTER experiments based on the lowest pretraining validation loss (see Supplementary Figure S2 for pretraining loss curves across text encoder variants). Cached embeddings were projected to the backbone hidden dimension using a two-layer multilayer perceptron with GELU activation.

\subsubsection{Numeric Pathway via FiLM}

For events with numeric values, PORTER constructs a fixed feature vector capturing measurement magnitude and, when available, relative abnormality without per-concept population statistics. When a valid reference range was available, the numeric scalar encoded the value's position within the range, with 0 corresponding to the lower limit and 1 to the upper limit, clipped to {[}-2, 3{]}. These asymmetric bounds reflect the heavier upper tail of clinical lab values while retaining at least 99\% of SickKids training data. When no reference range was available, PORTER used a log-magnitude fallback, log(1 + \textbar v\textbar) / 7, clipped to {[}0, 1.5{]}. The divisor of 7 maps log(1 + 1000) $\approx$ 6.9 to approximately 1.0. In SickKids training data, the 99th and 99.9th percentiles of the resulting pre-clip scalar were 0.83 and 1.32, respectively. Supplementary Figure S3 provides the empirical distribution of pre-clip numeric scalar features.

The numeric feature vector contained this scalar, indicators for reference-range scaling and log-magnitude fallback, and an indicator for numeric value presence. Exactly one scaling indicator was active when a numeric value was present. For events without numeric content, all indicators and the scalar were set to zero. This design preserves a fixed numeric feature layout across institutions and vocabularies without relying on concept-specific distributional summaries, such as means, standard deviations, or quantiles.

The feature vector was passed through a small multilayer perceptron with two 128-dimensional fully connected layers and GELU activations. Two linear heads then produced a residual scaling term $\gamma$ and a shift term $\beta$, each with dimension equal to the transformer hidden size. FiLM was applied to the projected text embedding h\textsubscript{text} as:

\[ h_{\text{fused}} = (1 + \gamma) \odot h_{\text{text}} + \beta \]
The $\gamma$ and $\beta$ heads were initialized to zero so that the fused representation was initially equal to the text representation. For events without numeric content, the numeric feature vector was all zeros, and the pathway was gated off by forcing $\gamma$ = 0 and $\beta$ = 0, yielding h\textsubscript{fused} = h\textsubscript{text}.

\subsubsection{Backbone Transformer}

The backbone was adapted from the architecture and hyperparameters used in our prior study\cite{RN13}. It was a decoder-only transformer\cite{RN34,RN35} with 28 layers, a hidden dimension of 768, and 12 attention heads, yielding 115.6 million backbone parameters. Temporal information was encoded using rotary position embeddings (RoPE)\cite{RN36} applied to patient age in days, so relative positional rotations reflected elapsed time between events. Transformer layers alternated between global attention and local attention\cite{RN37} with a window size of 128 tokens. Each clinical event contributed one position to the input sequence, corresponding to the event representation passed to the backbone. For numeric events, this representation was the text-derived event representation modulated by the numeric pathway, whereas for non-numeric events, it was the projected text-derived representation. Training batches were constructed using event-budget-based sequence packing, with a fixed budget of 16,384 clinical events per batch. Longer patient timelines were truncated, while shorter timelines were packed together using causal masking and patient-boundary masking to prevent cross-patient attention. The maximum effective context window was therefore 16,384 clinical events per patient. The model was implemented in PyTorch 2.7.

\subsubsection{Self-Supervised Pretraining}

PORTER was pretrained on the SickKids training split using autoregressive next-event prediction, implemented as classification of the next event into a finite output vocabulary of composite discrete tokens observed during pretraining. Following our prior work\cite{RN13}, each composite token combined the primary OMOP concept ID with discretized event attributes, including numeric values discretized into 10 quantile bins where applicable. In PORTER, these composite tokens served only as output labels for autoregressive pretraining, while model inputs were language-grounded event representations. The output head was therefore discarded before downstream representation extraction. Thus, downstream portability depends on the frozen input pathway and temporal backbone, not on reuse of the pretraining output vocabulary. Training used AdamW\cite{RN38} with a peak learning rate of 5$\times$10\textsuperscript{-4}, cosine decay with warmup, weight decay 0.05, gradient clipping at 1.0, Adam $\beta$\textsubscript{1} = 0.9, Adam $\beta$\textsubscript{2} = 0.95, and bfloat16 mixed precision. Models were trained for five epochs on a single NVIDIA H100 or L40S GPU without early stopping.\cite{RN39} Complete optimizer and training hyperparameters are provided in Supplementary Table S4.

\subsection{Comparison Models}

We compared PORTER against two comparators: a fixed-vocabulary foundation model representing the standard discrete-token input approach\cite{RN1,RN2}, and a patient-level text serialization comparator representing task-specific text serialization.

The Fixed-Vocab FM served as a controlled comparator that differed from PORTER in how clinical events were represented at input. Unlike PORTER, where composite tokens served only as output labels during pretraining, Fixed-Vocab FM used the same composite tokens as both input identifiers and output labels. Its input layer was therefore a learned embedding table indexed by composite token, and its prediction head predicted the next composite token from the finite pretraining vocabulary. Pretraining data, backbone transformer, temporal encoding, and training schedule were otherwise matched to PORTER. During transfer evaluation, Fixed-Vocab FM could only consume composite tokens observed during SickKids pretraining. Composite tokens outside this vocabulary were dropped before representation extraction.

The patient-level text serialization comparator followed a representative recent approach.\cite{RN15} For each task, the patient timeline up to the task-specific prediction time was serialized into a structured markdown document containing demographics, recent body metrics, recent vital signs, recent laboratory results, past visits, and visit-level summaries (see Supplementary Figure S4 for an illustrative serialized patient timeline). Recent measurements were drawn from curated code lists and limited to the most recent available values. As in the prior study, each document was truncated to 8,192 input tokens after prepending a task-specific retrieval instruction (see Supplementary Table S5 for the full set of task-specific instructions). The resulting text was encoded with Qwen3-Embedding-8B using last-token pooling to produce a single 4,096-dimensional task-conditioned patient representation. Because the retrieval instruction was task-specific, patient-level embeddings were generated separately for each task. In contrast, PORTER and Fixed-Vocab FM produced task-agnostic patient representations that were reused across downstream tasks. We therefore used the patient-level text serialization comparator only in in-domain evaluation, not in cross-vocabulary or cross-site evaluations.

\subsection{Ablation Experiments}

We conducted two sets of ablation experiments to evaluate key architectural decisions, with all variants pretrained on the same SickKids training split using the same backbone, pretraining objective, optimizer, and training schedule.

\subsubsection{Text Encoder Ablation}

We compared three frozen text encoders: BioLORD-2023 (768-dimensional), BGE-M3 (1,024-dimensional), and Qwen3-Embedding-8B (4,096-dimensional), plus a non-semantic random-embedding baseline. For the random baseline, each unique rendered event-description string was assigned a fixed 4,096-dimensional vector drawn from a standard normal distribution and L2-normalized to unit length. Vectors were deterministically keyed by the rendered event-description string, so identical event-description strings received identical vectors across vocabularies, whereas semantically equivalent but textually distinct event-description strings received uncorrelated vectors. This baseline preserves event-description identity while removing semantic structure. The lookup table was cached in the same manner as the text-encoder embeddings. Because embedding dimensionality differed across encoders, the input dimension of the text projection MLP (and consequently its parameter count) also varied, while the templated event descriptions and numeric pathway remained identical.

\subsubsection{Numeric Encoding Ablation}

We compared three architectural strategies for integrating numeric values into the language-grounded event representation: 1) PORTER, which encodes numeric values through the numeric pathway and integrates them with the text-derived event representation through FiLM; 2) PORTER-NoNum, which removes the numeric pathway and FiLM entirely and uses only the text-derived representation; and 3) PORTER-NumText, which removes the numeric pathway and FiLM but incorporates numeric values, units, and reference ranges directly into the templated event descriptions before encoding, testing whether the frozen text encoder alone could represent numeric magnitude when numeric information was expressed in language.

\subsection{Downstream Evaluation}

\subsubsection{Full-Shot Evaluation}

After pretraining, the backbone was frozen and hidden states were extracted at prediction time for each of the 74 downstream binary classification tasks. Linear probes were trained using L2-regularized logistic regression implemented in Scikit-learn\cite{RN40}, with features standardized to zero mean and unit variance using statistics computed on the training subset of each task\textquotesingle s split. Models were fit with LBFGS for up to 10,000 iterations, with inverse regularization strength selected from \{1, 10\textsuperscript{-1}, 10\textsuperscript{-2}, 10\textsuperscript{-3}, 10\textsuperscript{-4}\} on the validation set. Performance was reported as AUROC on the held-out temporal test split.

\subsubsection{Sample Efficiency Evaluation}

To assess sample efficiency, models were evaluated across labeled training set sizes (k) ranging from 2 to 32,768 examples, in powers of two. For each shot size, 10 independent iterations were conducted using a fixed sequence of random seeds applied identically across tasks. In each iteration, k training examples were drawn from the task-specific training set using balanced sampling (equal positive and negative examples where possible), while the validation and test sets remained unchanged. For tasks with fewer than k/2 positive instances available, all available positive instances were included, with the remaining examples drawn from the negative class. The same linear probe protocol was used. Task performance was reported as the mean across the 10 iterations for each shot size.

\subsubsection{Cross-Vocabulary Evaluation}

To isolate vocabulary shift from differences in patient population, outcome definitions, and temporal evaluation period, the cross-vocabulary evaluation used the same underlying SickKids admissions, prediction times, labels, and downstream splits as the in-domain evaluation. Only the source naming system used to render event descriptions was changed. In the primary PORTER setup, OMOP-derived event descriptions were constructed from OMOP concept names. In the cross-vocabulary setup, native SEDAR-derived event descriptions for the same patient timelines were regenerated using SEDAR concept names. For PORTER, the cached text-embedding table was regenerated from these alternate event descriptions using the same frozen text encoder, while the projection layer, FiLM module, and temporal backbone remained unchanged and frozen. This tested whether a PORTER model pretrained with OMOP-derived descriptions could consume the same clinical timelines rendered using a source naming system not seen during pretraining, without retraining the input pathway or backbone. Fixed-Vocab FM could not be evaluated in this setting because its input embedding table was indexed by composite event tokens observed during pretraining, and SEDAR composite event tokens had no learned input embeddings. Downstream evaluation followed the same linear-probe full-shot and sample-efficiency evaluation protocols used for in-domain evaluation.

To provide an upper-bound reference, we trained a Fixed-Vocab FM from scratch on SickKids data using composite event tokens constructed from native SEDAR event-attribute combinations, and evaluated it under the same full-shot protocol.

\subsubsection{Cross-Site Evaluation}

Cross-site generalization was evaluated by applying models pretrained on SickKids to MIMIC. For PORTER, MIMIC OMOP-derived event descriptions were rendered using the same templates, with slots populated from MIMIC OMOP concept names, then encoded with the same frozen text encoder to construct a MIMIC-specific cached embedding table. The SickKids-trained projection layer, FiLM module, and temporal backbone were reused unchanged. Thus, PORTER adapted its input cache to the MIMIC OMOP-derived description set through text encoding, without modifying learned model weights before downstream evaluation.

Fixed-Vocab FM was transferred by applying its SickKids-trained composite-token vocabulary to MIMIC composite tokens. MIMIC events whose composite token was not observed during SickKids pretraining were dropped before representation extraction. We quantified event-drop rates as the fraction of MIMIC events removed by this procedure. In contrast, PORTER does not drop any events: any MIMIC event whose event description the frozen text encoder can read produces a representation.

Downstream evaluation for PORTER and Fixed-Vocab FM followed the same linear-probe full-shot and sample-efficiency evaluation protocols used for in-domain evaluation.

To provide an upper-bound reference, we trained a Fixed-Vocab FM from scratch on MIMIC data using the same pretraining procedure and evaluated it under the full-shot protocol.

\subsection{Mechanistic Analyses}

We performed representation-level analyses to characterize mechanisms underlying cross-vocabulary transfer and numeric encoding. To assess whether cross-vocabulary transfer preserved patient-level geometry, we extracted test-set patient representations from PORTER variants with different text encoders after rendering the same patient timelines under two source naming systems: OMOP-derived descriptions and native SEDAR-derived descriptions. The conditions differed only in the names used to construct event descriptions, whereas the underlying events, timestamps, and numeric metadata were identical. We computed pairwise cosine distances among test patients under each condition. Relative geometry was quantified as the Spearman correlation between the OMOP-derived and SEDAR-derived distance matrices. Preservation of distance scale was summarized by ordinary least-squares regression of SEDAR-derived distances on OMOP-derived distances, with slope near 1 and intercept near 0 indicating minimal compression, expansion, or shift across naming systems.

At the event level, we evaluated synonym invariance and numeric sensitivity of the event representations passed to the transformer for PORTER, PORTER-NoNum, and PORTER-NumText. Synonym invariance was assessed using OMOP-derived and SEDAR-derived event-description pairs across drug, measurement, procedure, condition, and observation events. Because multiple SEDAR concept names could correspond to the same OMOP concept, this analysis tested whether SEDAR-derived naming variants were mapped near their shared OMOP-derived reference. We retained concepts with at least two distinct SEDAR-derived event descriptions and excluded queries in which the OMOP concept name appeared within the SEDAR-derived description. For each concept, up to 10 SEDAR-derived descriptions were used as queries. The matched OMOP-derived description served as the within-concept reference, and 32 sampled OMOP-derived descriptions from other concepts in the same domain served as distractors. Cosine similarity was compared between query-reference and query-distractor pairs.

Numeric sensitivity was assessed among measurement concepts with at least 1,000 numeric values recorded in a common unit. For each concept, we constructed the event representation passed to the transformer at empirical deciles of the observed value distribution for PORTER, PORTER-NoNum, and PORTER-NumText. Representations were L2-normalized. For each concept, cosine distance from the first-decile representation to each subsequent decile representation was computed, and Spearman correlation between decile rank and distance was measured. Higher positive correlation indicated stronger monotonic sensitivity of the representation to numeric magnitude. PORTER-NoNum was invariant to decile by construction, so its correlation was undefined.

\subsection{Compute Cost Analysis}

Total compute for PORTER was estimated as the sum of backbone pretraining floating-point operations (FLOPs), one-time event description text-embedding FLOPs, and backbone extraction FLOPs for generating patient representations over the downstream task cohorts. Backbone pretraining was estimated using the standard 6ND\cite{RN41} approximation, where N is the number of non-embedding backbone parameters and D is the number of EHR event tokens processed during training. Event description text embeddings were computed once over unique event descriptions and estimated as 2N\textsubscript{text}D\textsubscript{text} forward-pass FLOPs through BioLORD, where N\textsubscript{text} is the number of non-embedding text-encoder parameters and D\textsubscript{text} is the total number of event-description tokens. Backbone extraction was estimated as the forward-pass FLOPs of the pretrained EHR backbone over patient records in the task cohort. For Fixed-Vocab FM, only backbone pretraining and backbone extraction FLOPs applied.

For the text serialization comparator, per-task compute was estimated as Qwen3-Embedding-8B forward-pass FLOPs summed over all patient records in that task's cohort. Because the serialized input and retrieval instruction were task-specific, representations could not be shared across tasks, so total compute scaled with the number of downstream tasks. We report aggregate compute across the 74 SickKids tasks and its effective scaling factor relative to PORTER.

\subsection{Statistical Analysis}

Model comparisons were based on paired per-task AUROC values. For each model pair, we computed per-task AUROC differences across the 74 SickKids tasks for in-domain and cross-vocabulary evaluations, or the 36 MIMIC tasks for cross-site evaluation, and tested whether the paired differences were symmetrically centered around zero using a two-sided Wilcoxon signed-rank test\cite{RN42} implemented in SciPy.\cite{RN43} When multiple comparisons were performed within an ablation family (text encoder choice or numeric encoding strategy), p-values were adjusted using Holm\textquotesingle s step-down procedure \cite{RN44} to control the family-wise error rate at $\alpha$ = 0.05. Holm correction was applied separately within each evaluation setting: in-domain, cross-vocabulary, and cross-site.

\section{RESULTS}

\subsection{Study Cohorts, Vocabularies, and Pretraining Compute}

The SickKids and MIMIC pretraining cohorts comprised 2,100,646 and 339,989 patients, with 220.3 and 181.3 million clinical events, respectively (Table 1). Downstream evaluation included 101,404 SickKids admissions and 58,513 MIMIC admissions. The SickKids OMOP, SEDAR, and MIMIC OMOP source concept sets included 15,516, 105,551, and 32,032 unique clinical concepts, with 19,686, 523,724, and 36,524 unique rendered event-descriptions, respectively. SickKids OMOP and MIMIC OMOP shared 1,153 event-descriptions, whereas SickKids OMOP and SEDAR shared only 2. Across models, pretraining compute was similar at approximately 1.0 $\times$ 10\textsuperscript{18} FLOPs. Trainable input-pathway parameters ranged from 1.18M for PORTER variants without numeric integration to 75.5M for the Fixed-Vocab FM learned vocabulary embedding table (Supplementary Table S6).

\begin{table}[ht]
\centering
\caption{Characteristics of pretraining and task cohorts.}
\label{tab:cohorts}
\begin{tabular}{lcc}
\toprule
\textbf{Characteristic} & \textbf{SickKids} & \textbf{MIMIC} \\
\midrule
\multicolumn{3}{l}{\textit{Pretraining Cohort}\textsuperscript{a}}\\
\quad Patients, n & 2,100,646 & 339,989 \\
\quad Clinical events, n & 220,296,987 & 181,321,218 \\
\quad Timeline duration in days, median (IQR) & 1 (0--256) & 17 (2--553) \\
\quad Female sex, \% & 49.7 & 53.5 \\
\addlinespace
\multicolumn{3}{l}{\textit{Task Cohort}\textsuperscript{b}}\\
\quad Patients, n & 58,426 & 44,055 \\
\quad Admissions, n & 101,404 & 58,513 \\
\quad Age at admission, median (IQR) & 7 (2--13) & 54 (34--70) \\
\quad Length of stay in days, median (IQR) & 2 (1--6) & 4 (2--7) \\
\quad Female sex, \% & 45.9 & 61.9 \\
\bottomrule
\end{tabular}

\vspace{4pt}
{\footnotesize\begin{flushleft}
\textsuperscript{a}Pretraining cohorts include patients and events used for self-supervised pretraining of the SickKids and the reference MIMIC foundation models.\\
\textsuperscript{b}Task cohorts include admissions considered for downstream clinical prediction evaluation. The MIMIC cohort was also used for the external evaluation of the SickKids foundation models.\\
Abbreviations: IQR, interquartile range; SickKids, The Hospital for Sick Children; MIMIC, Medical Information Mart for Intensive Care.
\end{flushleft}}
\end{table}

\subsection{PORTER Matches Fixed-Vocab FM In-Domain and Improves Transfer Across Vocabularies and Institutions}

On the 74 SickKids in-domain tasks, PORTER and Fixed-Vocab FM did not differ in mean AUROC (0.884 vs 0.884, p=0.942; Table 2, Supplementary Table S7). In the cross-vocabulary setting, SickKids timelines were re-expressed using SEDAR-derived event descriptions. PORTER achieved a mean AUROC of 0.848 without retraining or vocabulary mapping, recovering 97.1\% of the performance of an upper-bound reference model trained directly on SEDAR composite event tokens (mean AUROC, 0.873). Fixed-Vocab FM could not be directly evaluated in this setting because its learned input vocabulary did not contain SEDAR composite event tokens. On cross-site evaluation in MIMIC, PORTER had higher AUROC than Fixed-Vocab FM on 31 of 36 tasks (mean AUROC 0.823 vs 0.810, p\textless0.001). For Fixed-Vocab FM, 69\% of MIMIC events contained unseen composite tokens and were dropped during representation extraction, whereas PORTER produced representations for all MIMIC events. An upper-bound reference model trained directly on MIMIC achieved mean AUROC of 0.848. Across labeled training-set sizes, PORTER and Fixed-Vocab FM were similar in-domain at larger sample sizes, while PORTER was higher in few-shot in-domain settings and across sample sizes in MIMIC (Supplementary Figure S5).

\begin{table}[ht]
\centering
\caption{In-domain, cross-vocabulary, and cross-site mean AUROC of PORTER and Fixed-Vocab FM.}
\label{tab:main-auroc}
\begin{tabular}{lccc}
\toprule
 & In-domain (SK) & Cross-vocabulary (SK) & Cross-site (MIMIC) \\
\midrule
\textbf{PORTER} & \textbf{0.884} & \textbf{0.848} & \textbf{0.823} \\
Fixed-Vocab FM & 0.884 & --\textsuperscript{a} & 0.810\textsuperscript{***} \\
\bottomrule
\end{tabular}

\vspace{4pt}
{\footnotesize\begin{flushleft}
Bold: the statistical reference model.\\
\textsuperscript{***}$p<0.001$ vs PORTER (reference) using two-sided Wilcoxon signed-rank test.\\
\textsuperscript{a}Fixed-Vocab FM was not evaluated in cross-vocabulary transfer because SEDAR composite event tokens had no learned input embeddings.\\
Abbreviations: AUROC, area under the receiver operating characteristic curve; PORTER, Portable EHR Representations; EHR, electronic health records; FM, foundation model; SK, SickKids; MIMIC, Medical Information Mart for Intensive Care; SEDAR, SickKids Enterprise-wide Data in Azure Repository.
\end{flushleft}}
\end{table}

\subsection{PORTER Outperforms a Task-Specific Text Serialization Comparator on AUROC, Sample Efficiency, and Amortized Compute}

PORTER had higher full-shot AUROC than the patient-level text serialization comparator on 69 of 74 in-domain tasks (mean AUROC 0.884 vs 0.863, p\textless0.001; Figure 2A, Supplementary Table S8). Across labeled training-set sizes, PORTER had higher AUROC from 4 examples per task onward, with larger differences between 32 to 512 examples per task (Figure 2B). The text serialization comparator required less compute when only a small number of patient-task representations were generated, whereas PORTER required less compute once representations were generated for more than approximately 256 admissions per task, and required 329-fold fewer amortized FLOPs across all 74 task cohorts (Figure 2C).

\begin{figure}[t]
\centering
\includegraphics[width=\linewidth]{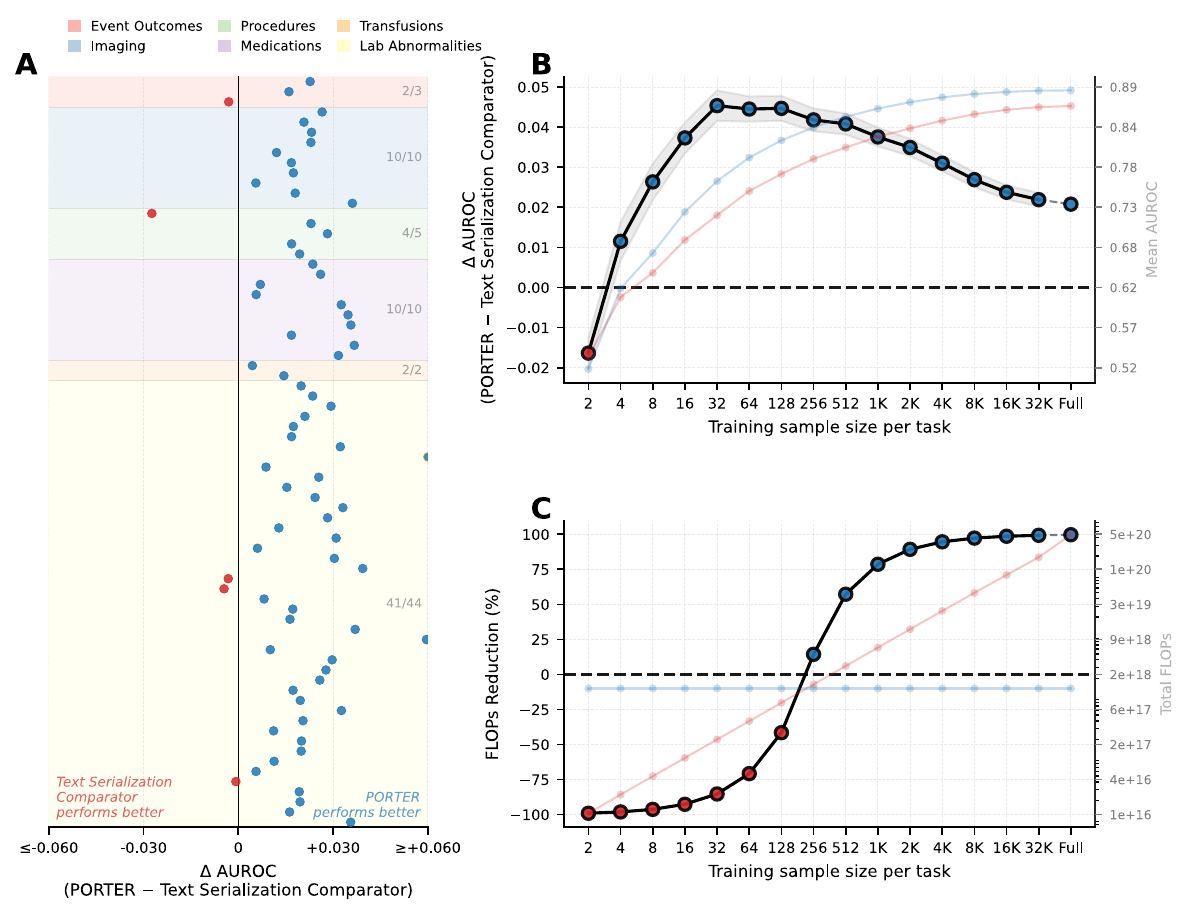}
\caption{Performance, sample efficiency, and compute for PORTER versus the patient-level text serialization comparator. (A) Per-task AUROC difference between PORTER and the patient-level text serialization comparator (Qwen3-Embedding-8B with patient-level text serialization) across 74 SickKids clinical prediction tasks evaluated in-domain. Each point represents one task. Blue points indicate tasks where PORTER performed better, and red points indicate tasks where the text serialization comparator performed better. Tasks are grouped by clinical category, with per-category win counts annotated. (B) Mean AUROC difference between PORTER and the text serialization comparator, left axis, and mean AUROC for each approach, right axis, as a function of labeled training examples per task. Shaded region indicates $\pm 1$ standard error of the mean difference across tasks. (C) Percentage reduction in total floating-point operations achieved by PORTER relative to the text serialization comparator, left axis, alongside total FLOPs for each approach, right axis. PORTER patient representations are computed once and reused across tasks, whereas the text serialization comparator requires a separate forward pass through Qwen3-Embedding-8B for each patient-task pair. As a result, comparator compute scales with the number of downstream tasks, while PORTER's amortized compute advantage increases with downstream reuse. Abbreviations: PORTER, Portable EHR Representations; EHR, electronic health records; AUROC, area under the receiver operating characteristic curve; FLOPs, floating-point operations.}
\label{fig:performance}
\end{figure}

\subsection{Text Encoder Choice Determines Cross-Vocabulary Transfer More Than In-Domain Performance}

Holding the numeric pathway and FiLM fixed, encoder-related AUROC differences were largest in the cross-vocabulary setting, where SickKids OMOP and SEDAR shared almost no event-description strings (Table 3, Supplementary Table S9). In-domain differences were statistically significant but small: 0.884 for BioLORD, 0.881 for Qwen3, 0.881 for BGE-M3, and 0.879 for Random (all p\textless0.001 vs BioLORD). Cross-vocabulary differences were larger: 0.848, 0.840, 0.818, and 0.780, respectively, with the largest degradation observed for Random (all p\textless0.001 vs BioLORD). On cross-site evaluation, Qwen3 (0.825) and BGE-M3 (0.821) were statistically indistinguishable from BioLORD (0.823), while Random was lower (0.812; p\textless0.001). Sample-efficiency curves showed limited in-domain differences but larger cross-vocabulary and cross-site performance deficits relative to BioLORD across labeled sample sizes, again most prominently for Random (Supplementary Figure S6).

\begin{table}[ht]
\centering
\caption{Effect of text encoder choice on in-domain, cross-vocabulary, and cross-site mean AUROC.}
\label{tab:encoder}
\setlength{\tabcolsep}{4pt}
\begin{tabular}{llccc}
\toprule
Input representation & Text Encoder Parameters & In-domain (SK) & Cross-vocabulary (SK) & Cross-site (MIMIC) \\
\midrule
\textbf{BioLORD} & \textbf{0.1B} & \textbf{0.884} & \textbf{0.848} & \textbf{0.823} \\
Qwen3 & 8B & 0.881\textsuperscript{***} & 0.840\textsuperscript{***} & 0.825 \\
BGE-M3 & 0.6B & 0.881\textsuperscript{***} & 0.818\textsuperscript{***} & 0.821 \\
Random Embeddings & -- & 0.879\textsuperscript{***} & 0.780\textsuperscript{***} & 0.812\textsuperscript{***} \\
\bottomrule
\end{tabular}

\vspace{4pt}
{\footnotesize\begin{flushleft}
Bold: the statistical reference model.\\
Significance vs BioLORD (reference); \textsuperscript{***}$p<0.001$ (two-sided Wilcoxon signed-rank test, Holm-corrected across 3 reference comparisons per setting).\\
Abbreviations: AUROC, area under the receiver operating characteristic curve; SK, SickKids; MIMIC, Medical Information Mart for Intensive Care; B, billion.
\end{flushleft}}
\end{table}

Preservation of patient-level representation geometry between OMOP-derived and SEDAR-derived event descriptions followed the same across-encoder ranking as cross-vocabulary AUROC. Pairwise cosine distances between test patients computed under the two description sources had Spearman correlations of 0.613 for BioLORD, 0.606 for Qwen3, 0.515 for BGE-M3, and 0.267 for Random. Distance-scale preservation showed the same ranking, with OLS slopes/intercepts of 0.660/0.018, 0.602/0.086, 0.478/0.137, and 0.318/0.210, respectively (Supplementary Figure S7).

\subsection{FiLM Numeric Pathway Improves Numeric Sensitivity Without Disrupting Concept Identity}

Holding the text encoder fixed, mean AUROC differed by numeric encoding strategy (Table 4, Supplementary Table S10). PORTER had the highest mean AUROC in all three settings: 0.884 in-domain versus 0.880 for PORTER-NumText and 0.873 for PORTER-NoNum; 0.848 cross-vocabulary versus 0.839 and 0.837; and 0.823 cross-site versus 0.801 and 0.804, respectively (all p\textless0.001 vs PORTER). Sample-efficiency curves followed the same overall pattern (Supplementary Figure S8).

\begin{table}[ht]
\centering
\caption{Effect of numeric encoding strategy on in-domain, cross-vocabulary, and cross-site mean AUROC.}
\label{tab:numeric}
\begin{tabular}{lccc}
\toprule
Variant & In-domain (SK) & Cross-vocabulary (SK) & Cross-site (MIMIC) \\
\midrule
\textbf{PORTER} & \textbf{0.884} & \textbf{0.848} & \textbf{0.823} \\
PORTER-NoNum & 0.873\textsuperscript{***} & 0.837\textsuperscript{***} & 0.804\textsuperscript{***} \\
PORTER-NumText & 0.880\textsuperscript{***} & 0.839\textsuperscript{***} & 0.801\textsuperscript{***} \\
\bottomrule
\end{tabular}

\vspace{4pt}
{\footnotesize\begin{flushleft}
Bold: the statistical reference model.\\
Significance vs PORTER (reference); \textsuperscript{***}$p<0.001$ (two-sided Wilcoxon signed-rank test, Holm-corrected across 2 reference comparisons per setting).\\
Abbreviations: AUROC, area under the receiver operating characteristic curve; PORTER, Portable EHR Representations; EHR, electronic health records; NoNum, no numeric pathway (text-derived representations only); NumText, numeric values rendered as text; SK, SickKids; MIMIC, Medical Information Mart for Intensive Care. MIMIC cross-site results include 36 tasks.
\end{flushleft}}
\end{table}

Analyses of the event input representations passed to the transformer showed that all three variants preserved clinical concept identity, with within-concept cosine similarity exceeding across-concept similarity across event domains (Supplementary Figure S9). For numeric sensitivity, the Spearman correlation for each measurement concept between value-decile rank and cosine distance from the first-decile representation had median 1.00 for PORTER and 0.67 for PORTER-NumText. PORTER-NoNum was invariant to numeric value by construction, so this correlation was undefined.

\section{DISCUSSION}

PORTER enabled direct cross-vocabulary transfer to patient timelines expressed with event descriptions not used during pretraining, achieving 97.1\% of the AUROC of a reference model trained directly on the target vocabulary. This transfer required neither PORTER retraining nor vocabulary mapping. PORTER also matched a fixed-vocabulary foundation model across 74 in-domain tasks and improved transfer on 31 of 36 tasks at an external site. Compared with a task-specific patient-level text serialization comparator, PORTER achieved higher AUROC on 69 of 74 in-domain tasks and showed an amortized compute advantage that increased with the number of downstream tasks and patients, reaching 329-fold in the full evaluation. Ablations clarified the distinct roles of PORTER's architectural components. The temporal backbone preserved in-domain performance across input representations. Cross-vocabulary performance tracked how well each encoder preserved patient-level representation geometry across vocabularies, an emergent property of the pretrained representations that is not explained by encoder scale alone. The dedicated numeric pathway improved sensitivity to numeric magnitude without degrading clinical concept identity.

PORTER shows that vocabulary-independent inputs can support self-supervised EHR foundation model pretraining without compromising task-agnostic representation learning. Prior work has shown that encoding clinical concepts or events through text can reduce dependence on fixed vocabularies and improve transfer across institutions and languages.\cite{RN18,RN19,RN22,RN23} However, these approaches generally use text-based representations within supervised settings. Even when self-supervised objectives are used, the resulting model is typically adapted through downstream fine-tuning rather than used as a frozen source of task-agnostic patient representations. PORTER instead uses autoregressive pretraining to learn a temporal backbone that is frozen and reused across downstream tasks through linear probes. The cross-vocabulary evaluation isolates vocabulary shift from the population and practice differences that confound cross-site comparisons. This setting also reflects a common deployment challenge in which models trained on retrospective research datasets may later be applied to production feeds that use different vocabularies or naming conventions. Fixed-vocabulary foundation models lack a direct mechanism for this setting without vocabulary mapping or retraining. Together with the in-domain parity result, these findings support vocabulary-independent input representations as a viable design choice for structured EHR foundation models.

Ablation experiments clarified the distinct roles of PORTER's architectural components. Under in-domain evaluation, random embeddings preserved much of PORTER's AUROC. This indicates that the temporal backbone can learn useful clinical dynamics when clinical events are represented consistently, even without semantic structure in the input vectors. In cross-vocabulary evaluation, however, random embeddings degraded performance substantially, likely because distinct descriptions of the same clinical concept were assigned unrelated embeddings. Across PORTER variants using different text encoders, cross-vocabulary AUROC tracked the extent to which each encoder preserved patient-level pairwise distance structure across vocabularies. Next-event prediction does not optimize for this geometry, so this correspondence is an emergent property of the pretrained representations and points to semantic alignment as the basis for vocabulary transfer. Finally, using the dedicated numeric pathway increased ordinal sensitivity from 0.67 to 1.00 compared to text-encoded numeric values while preserving within-concept representation structure. This indicates that FiLM-based numeric integration strengthened numeric sensitivity in the event input representation without disrupting the concept semantics supplied by the event-description embedding. The advantage of the dedicated numeric pathway over rendering numeric values as text widened in cross-vocabulary and cross-site evaluations, consistent with normalized numeric features transferring more reliably than text-rendered magnitudes. In PORTER, this normalization is reference-range anchored when available and uses a bounded log-magnitude fallback otherwise. Together, these findings indicate that PORTER\textquotesingle s components serve distinct and complementary roles and can in principle be improved independently.

The encoder ablation findings suggest that text-encoder scale is not the primary determinant of PORTER's performance. Across evaluations, BioLORD, a 109M-parameter biomedical encoder trained to align biomedical concept names with ontology-informed definitions, performed on par with or better than general-purpose encoders up to 63 times larger, with the largest advantage in cross-vocabulary evaluation. This pattern may reflect PORTER's narrower requirement for the text encoder, which is to embed short clinical event descriptions so that alternative descriptions of the same clinical concepts remain close across vocabularies. By contrast, Qwen3-Embedding and BGE-M3 are general-purpose embedding models optimized across broad multilingual, retrieval, relevance-ranking, and long-context settings. Standard embedding benchmarks evaluate many useful capabilities\cite{RN45}, but they do not directly measure the properties most relevant to language-grounded EHR event modeling, including preservation of patient-level geometry across vocabularies and ordinal sensitivity to numeric values. Developing encoder benchmarks around these properties could make encoder selection more efficient and may improve cross-vocabulary transfer without changing the temporal backbone or numeric pathway.

The text encoder findings also inform the comparison between PORTER and patient-level text serialization. In PORTER, language grounding occurs at the event level. Each clinical event is represented using a frozen text encoder and dedicated numeric pathway, and the temporal backbone integrates sequences of event input representations into reusable patient representations. Patient-level serialization instead represents the full patient timeline as text and encodes it directly as a patient-level representation, often within a task-specific prompt. This provides flexible task conditioning but requires the model to represent the full task-conditioned patient timeline. Prior patient-level serialization work suggests that this setting favors larger general-purpose encoders.\cite{RN15} The compute profiles also differ. PORTER's event-description embeddings are computed once per unique description, and the resulting patient representations can be reused across downstream tasks. In contrast, the task-specific serialization comparator we evaluate requires a language-model forward pass for each patient-task pair. Recent rubric-based serialization methods synthesize an extraction schema from labeled examples and report higher accuracy than direct serialization\cite{RN14}, although they still build patient representations tied to a specific task. These differences suggest that PORTER suits high-throughput, population-scale prediction, whereas patient-level serialization may suit settings that value flexible, task-specific conditioning over reuse.

Several limitations should be noted. All models in the main experiments were pretrained on data from a single pediatric hospital, and cross-site evaluation was limited to one external adult intensive care setting, where vocabulary mismatch and population shift co-occur. All downstream evaluations used linear probes on frozen representations for binary classification tasks. Performance under fine-tuning, more expressive adaptation methods\cite{RN46,RN47}, regression tasks, zero-shot prediction, or generative settings has not been assessed. We also kept the text encoder frozen and did not test fine-tuning it end-to-end during pretraining, which could change which encoder properties matter.\cite{RN23} PORTER depends on the availability and quality of event descriptions and numeric metadata, including reference ranges, which may vary across institutions and data pipelines. Finally, although PORTER removes fixed-vocabulary dependence on the input side, its autoregressive pretraining objective still uses a fixed output vocabulary for next-event prediction. The language-grounded input pathway is independent of this objective and could be paired with alternatives such as time-to-event or continuous-time generative formulations.\cite{RN5,RN6,RN7,RN48} Because these alternatives are also typically defined over a fixed set of output codes, reducing output-side vocabulary dependence remains future work.

In conclusion, PORTER shows that language-grounded event representations can make structured EHR foundation models portable across vocabularies while preserving in-domain performance and allowing patient representations to be reused efficiently across downstream tasks. By modeling concept semantics, numeric values, and temporal dynamics separately, PORTER reduces dependence on fixed vocabularies and provides a practical route toward EHR foundation models that can generalize across deployment settings and institutions.

\section{Data Availability}

The SickKids dataset cannot be made publicly available due to patient privacy restrictions. Relevant data are available upon reasonable request to the corresponding author. The MIMIC dataset is publicly available through PhysioNet (https://physionet.org/content/mimiciv/1.0/) subject to credentialing and a data use agreement.

\section{Code Availability}

The codebase for EHR foundation model training will be made publicly available at \url{https://github.com/sungresearch/ehr-fm}.

\section{Acknowledgements}

LS is supported by the Canada Research Chair in Pediatric Oncology Supportive Care.

We thank Jason Alan Fries and Natalie Pageler for their helpful feedback on an earlier draft of this manuscript.

\section{Funding}

This research did not receive funding.

\section{Author Contribution}

L.L.G. conceptualized and designed the study with input from all authors. L.L.G. performed all experiments, analyzed and interpreted results with input from all authors. L.L.G. wrote the manuscript with input from all authors. L.L.G, A.P.Y, E.V, and L.S read and approved the final manuscript.

\section{Competing Interests}

The authors declare no competing interests.

\bibliographystyle{unsrtnat}
\bibliography{references}

\clearpage
\section*{Supplementary Material}
\renewcommand{\thefigure}{S\arabic{figure}}
\renewcommand{\thetable}{S\arabic{table}}
\setcounter{figure}{0}
\setcounter{table}{0}
\subsection*{Supplementary Table S1. Comparison of PORTER with related text-based approaches for structured EHR modeling}

\begingroup\scriptsize\setlength{\tabcolsep}{4pt}\begin{longtable}[]{@{}
  >{\raggedright\arraybackslash}p{(\columnwidth - 14\tabcolsep) * \real{0.0909}}
  >{\raggedright\arraybackslash}p{(\columnwidth - 14\tabcolsep) * \real{0.1380}}
  >{\raggedright\arraybackslash}p{(\columnwidth - 14\tabcolsep) * \real{0.1010}}
  >{\raggedright\arraybackslash}p{(\columnwidth - 14\tabcolsep) * \real{0.1515}}
  >{\raggedright\arraybackslash}p{(\columnwidth - 14\tabcolsep) * \real{0.1886}}
  >{\raggedright\arraybackslash}p{(\columnwidth - 14\tabcolsep) * \real{0.1279}}
  >{\raggedright\arraybackslash}p{(\columnwidth - 14\tabcolsep) * \real{0.0572}}
  >{\raggedright\arraybackslash}p{(\columnwidth - 14\tabcolsep) * \real{0.1380}}@{}}
\toprule\noalign{}
\endhead
\bottomrule\noalign{}
\endlastfoot
\textbf{Approach} & & \textbf{Methodological features} & & & & \textbf{Evaluation} & \\
& \textbf{Text representation granularity\textsuperscript{a}} & \textbf{Text encoder\textsuperscript{b}} & \textbf{EHR pretraining objective} & \textbf{Frozen patient representation reused across tasks} & \textbf{Explicit numeric value handling} & \textbf{Cross-site} & \textbf{Cross-vocabulary (same patient)\textsuperscript{c}} \\
Hegselmann et al.~{[}15{]} & Patient-level & Frozen (pretrained) & --- & --- & Partial & \checkmark & --- \\
LRRL {[}14{]} & Patient-level & Frozen (pretrained) & --- & --- & Partial & --- & --- \\
DescEmb {[}19{]} & Code- or Event-level & Train or fine-tune & --- & --- & Partial & \checkmark & --- \\
GenHPF {[}22{]} & Code- or Event-level & Train or fine-tune & Masked + contrastive & --- & Partial & \checkmark & --- \\
Hur et al.~2026 {[}23{]} & Code- or Event-level & Train or fine-tune & --- & --- & Partial & \checkmark & --- \\
GRASP {[}18{]} & Code- or Event-level & Frozen (pretrained) & --- & --- & --- & \checkmark & --- \\
GAME {[}20{]} & Code- or Event-level & Frozen (pretrained) & --- & --- & --- & \checkmark & --- \\
Attrach et al.~{[}21{]} & Code- or Event-level & Frozen (pretrained) & --- & --- & \checkmark & --- & --- \\
\textbf{PORTER} & \textbf{Code- or Event-level} & Frozen (pretrained) & \textbf{Autoregressive (next-event prediction)} & \textbf{\checkmark} & \textbf{\checkmark} & \textbf{\checkmark} & \textbf{\checkmark} \\
\end{longtable}\endgroup

Symbols. \checkmark indicates present; --- indicates absent or not applicable.

\textsuperscript{a} Patient-level serializes the whole record into a single text document. Code- or event-level renders each code or clinical event as text. For GAME, text is one of several inputs to its code embeddings.

\textsuperscript{b} Reflects each method\textquotesingle s primary or representative encoder configuration. Frozen (pretrained): a pretrained text encoder used without weight updates; train or fine-tune: encoder weights updated, from random or pretrained initialization.

\textsuperscript{c} The cross-vocabulary evaluation isolates vocabulary shift by holding patients, labels, and clinical events fixed and changing only the naming system

Abbreviations: EHR, electronic health record; FM, foundation model; LRRL, large language model rubric representation learning; GRASP, Generalizable Risk Assessment with Semantic Projection; GAME, graph alignment for multi-institutional EHR data; PORTER, Portable EHR Representations.

\clearpage
\subsection*{Supplementary Figure S1. Cohort construction for pretraining and downstream evaluation}

\includegraphics[width=\linewidth,height=0.85\textheight,keepaspectratio]{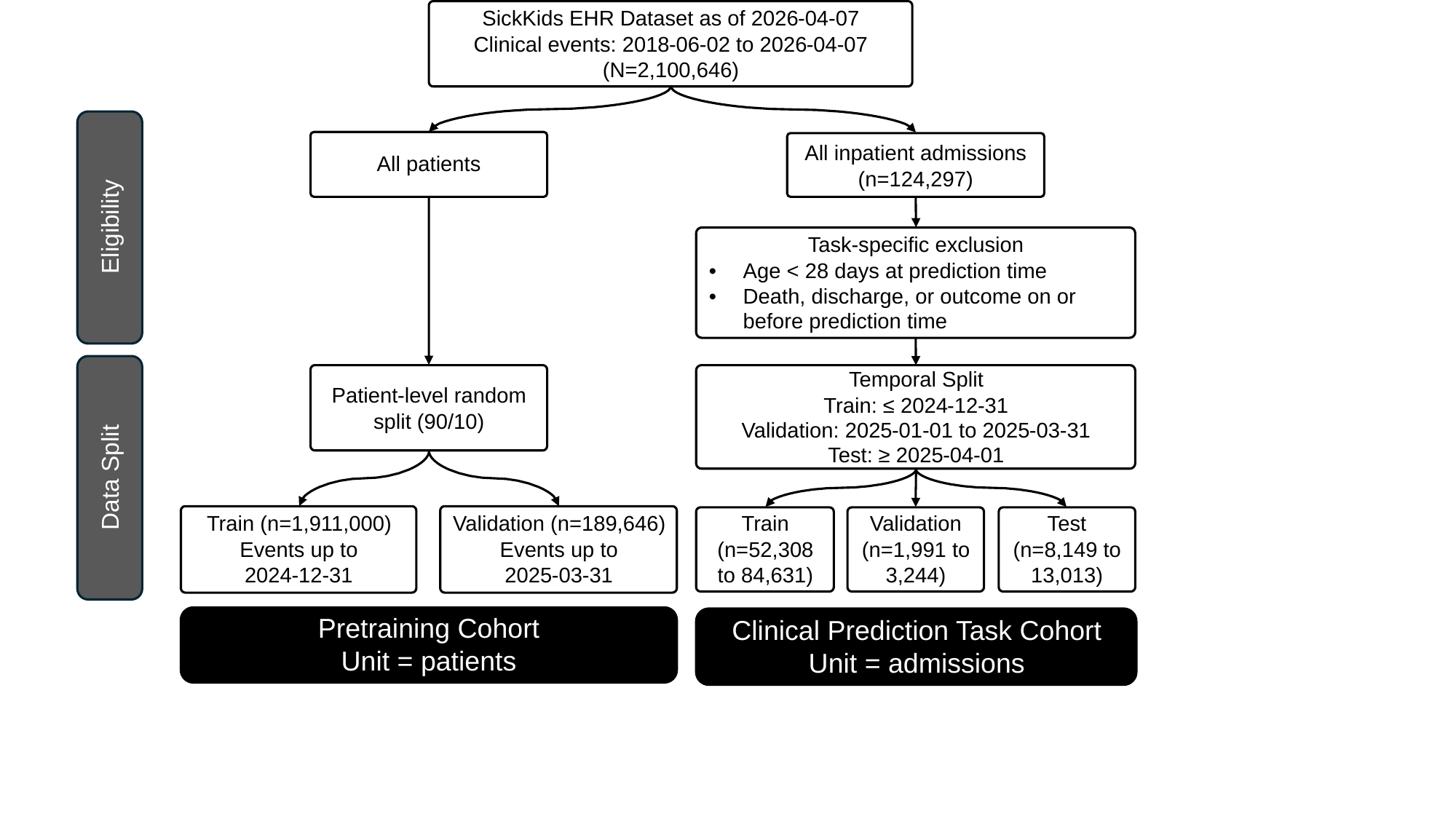}

Pretraining (left) used a patient-level random split, with different temporal cutoffs for training and validation to evaluate models on more recent clinical data during pretraining. Downstream evaluation (right) used admission-level temporal splits across 74 clinical prediction tasks, aligned to the same calendar cutoffs as pretraining. The unit of analysis was patients for pretraining and admissions for downstream prediction.

Abbreviations: EHR, electronic health records; SickKids, The Hospital for Sick Children.

\clearpage
\subsection*{Supplementary Table S2. SickKids and MIMIC task cohort statistics}

\begingroup\scriptsize\setlength{\tabcolsep}{4pt}\begin{longtable}[]{@{}
  >{\raggedright\arraybackslash}p{(\columnwidth - 8\tabcolsep) * \real{0.3016}}
  >{\raggedright\arraybackslash}p{(\columnwidth - 8\tabcolsep) * \real{0.1825}}
  >{\raggedright\arraybackslash}p{(\columnwidth - 8\tabcolsep) * \real{0.1667}}
  >{\raggedright\arraybackslash}p{(\columnwidth - 8\tabcolsep) * \real{0.1667}}
  >{\raggedright\arraybackslash}p{(\columnwidth - 8\tabcolsep) * \real{0.1667}}@{}}
\toprule\noalign{}
\begin{minipage}[b]{\linewidth}\raggedright
\textbf{Task}
\end{minipage} & \begin{minipage}[b]{\linewidth}\centering
\textbf{Total admissions}
\end{minipage} & \begin{minipage}[b]{\linewidth}\centering
\textbf{Total patients}
\end{minipage} & \begin{minipage}[b]{\linewidth}\centering
\textbf{Positive cases}
\end{minipage} & \begin{minipage}[b]{\linewidth}\centering
\textbf{Prevalence (\%)}
\end{minipage} \\
\midrule\noalign{}
\endhead
\bottomrule\noalign{}
\endlastfoot
\textbf{SickKids} & & & & \\
\textbf{Transfusions} & & & & \\
Platelet transfusion & 97,639 & 56,860 & 3,326 & 3.4 \\
Red cell transfusion & 92,090 & 54,100 & 7,367 & 8.0 \\
\textbf{Procedure} & & & & \\
Invasive intubation & 95,685 & 55,569 & 1,782 & 1.9 \\
Gastrostomy tube & 99,209 & 57,382 & 470 & 0.5 \\
Echocardiogram & 97,168 & 56,373 & 7,758 & 8.0 \\
Pulmonary function test & 99,223 & 57,383 & 848 & 0.9 \\
Lumbar puncture & 98,241 & 57,018 & 2,546 & 2.6 \\
Surgery & 76,757 & 43,698 & 13,623 & 17.7 \\
Interventional radiology & 98,251 & 57,179 & 8,309 & 8.5 \\
\textbf{Imaging} & & & & \\
Plain radiography chest & 88,431 & 52,133 & 10,411 & 11.8 \\
Ultrasound abdomen & 96,513 & 56,325 & 11,981 & 12.4 \\
Computerized tomography chest & 98,880 & 57,273 & 2,352 & 2.4 \\
Computerized tomography abdomen & 98,868 & 57,258 & 1,428 & 1.4 \\
Computerized tomography head & 97,799 & 56,743 & 3,398 & 3.5 \\
MRI head & 97,350 & 56,495 & 6,015 & 6.2 \\
MRI whole body & 99,249 & 57,382 & 181 & 0.2 \\
PET & 99,254 & 57,386 & 223 & 0.2 \\
\textbf{Laboratory abnormality} & & & & \\
High white blood count & 90,111 & 53,055 & 12,829 & 14.2 \\
Low white blood count & 95,165 & 55,980 & 9,835 & 10.3 \\
High absolute neutrophil count & 91,485 & 53,617 & 11,348 & 12.4 \\
Low absolute neutrophil count & 97,791 & 56,927 & 7,393 & 7.6 \\
High bands & 93,615 & 54,670 & 13,288 & 14.2 \\
High lymphocyte & 97,525 & 56,694 & 4,664 & 4.8 \\
Low lymphocyte & 92,867 & 54,773 & 13,948 & 15.0 \\
High hemoglobin & 95,409 & 55,740 & 4,335 & 4.5 \\
Low hemoglobin & 86,503 & 51,946 & 19,571 & 22.6 \\
High mean corpuscular volume & 95,707 & 56,292 & 6,733 & 7.0 \\
Low mean corpuscular volume & 95,576 & 55,804 & 5,066 & 5.3 \\
High reticulocyte count & 98,114 & 57,037 & 4,256 & 4.3 \\
Low reticulocyte count & 98,550 & 57,049 & 3,257 & 3.3 \\
High platelet & 94,099 & 55,398 & 12,402 & 13.2 \\
Low platelet & 90,823 & 53,949 & 13,118 & 14.4 \\
High immature platelet fraction & 97,245 & 56,684 & 5,804 & 6.0 \\
Low immature platelet fraction & 98,512 & 57,047 & 2,744 & 2.8 \\
High mean platelet volume & 97,647 & 56,802 & 4,904 & 5.0 \\
Low mean platelet volume & 94,819 & 55,317 & 8,600 & 9.1 \\
High fibrinogen & 98,626 & 57,033 & 2,357 & 2.4 \\
Low fibrinogen & 96,220 & 55,753 & 1,821 & 1.9 \\
High partial thromboplastin time & 96,050 & 55,989 & 3,734 & 3.9 \\
High international normalized ratio & 92,243 & 54,145 & 7,322 & 7.9 \\
High sodium & 91,134 & 53,508 & 10,158 & 11.1 \\
Low sodium & 95,319 & 55,879 & 8,529 & 8.9 \\
High potassium & 94,304 & 55,228 & 8,103 & 8.6 \\
Low potassium & 90,253 & 53,174 & 14,367 & 15.9 \\
High glucose & 86,739 & 51,197 & 12,280 & 14.2 \\
Low glucose & 97,176 & 56,699 & 4,174 & 4.3 \\
High creatinine & 95,557 & 55,956 & 5,327 & 5.6 \\
High urea & 97,735 & 56,967 & 3,579 & 3.7 \\
Low albumin & 95,514 & 55,840 & 11,950 & 12.5 \\
High alanine transaminase & 94,750 & 55,709 & 9,235 & 9.7 \\
High aspartate aminotransferase & 95,706 & 56,008 & 7,727 & 8.1 \\
High lactate dehydrogenase & 98,306 & 56,978 & 2,653 & 2.7 \\
High bilirubin & 94,637 & 55,260 & 5,682 & 6.0 \\
High cholesterol & 99,091 & 57,330 & 596 & 0.6 \\
High triglyceride & 98,593 & 57,026 & 2,527 & 2.6 \\
High ferritin & 98,066 & 56,786 & 3,542 & 3.6 \\
High creatinine kinase & 98,765 & 57,089 & 859 & 0.9 \\
High C-reactive protein & 92,870 & 54,447 & 12,666 & 13.6 \\
High erythrocyte sedimentation rate & 98,189 & 56,872 & 3,051 & 3.1 \\
Low PaO2 & 98,394 & 56,970 & 2,829 & 2.9 \\
Low SpO2 & 67,901 & 40,398 & 23,556 & 34.7 \\
\textbf{Medications} & & & & \\
Any antibacterial & 62,448 & 39,290 & 19,768 & 31.7 \\
Any antifungal & 98,452 & 57,331 & 2,424 & 2.5 \\
Any chemotherapy & 94,604 & 57,184 & 3,460 & 3.7 \\
Any antiepileptics & 88,672 & 54,112 & 7,545 & 8.5 \\
Any glucocorticoid & 76,125 & 45,350 & 15,972 & 21.0 \\
Dexamethasone & 79,737 & 46,577 & 11,147 & 14.0 \\
Any opioid & 67,830 & 39,184 & 18,109 & 26.7 \\
Morphine & 76,869 & 44,403 & 13,733 & 17.9 \\
Fentanyl & 79,101 & 45,636 & 13,959 & 17.6 \\
Any inotrope & 95,062 & 55,180 & 2,154 & 2.3 \\
\textbf{Clinical outcomes} & & & & \\
Long length of stay ($\geq$ 7 days) & 99,366 & 57,437 & 22,046 & 22.2 \\
Readmission within 30 days & 100,149 & 57,767 & 17,168 & 17.1 \\
Mortality & 99,258 & 57,387 & 580 & 0.6 \\
\textbf{MIMIC} & & & & \\
\textbf{Laboratory abnormality} & & & & \\
High white blood count & 44,613 & 34,696 & 12,036 & 27.0 \\
Low white blood count & 57,091 & 43,323 & 4,090 & 7.2 \\
High absolute neutrophil count & 54,519 & 41,165 & 3,470 & 6.4 \\
Low absolute neutrophil count & 58,321 & 43,961 & 588 & 1.0 \\
High lymphocyte & 58,311 & 43,924 & 327 & 0.6 \\
Low lymphocyte & 55,002 & 41,888 & 2,684 & 4.9 \\
Low hemoglobin & 35,499 & 28,992 & 15,397 & 43.4 \\
High mean corpuscular volume & 54,447 & 41,392 & 4,273 & 7.8 \\
Low mean corpuscular volume & 54,403 & 41,587 & 2,392 & 4.4 \\
High platelet & 56,924 & 43,179 & 3,169 & 5.6 \\
Low platelet & 52,218 & 39,892 & 8,889 & 17.0 \\
High fibrinogen & 57,321 & 43,351 & 2,205 & 3.8 \\
Low fibrinogen & 57,967 & 43,617 & 1,304 & 2.2 \\
High partial thromboplastin time & 52,680 & 40,039 & 8,128 & 15.4 \\
High international normalized ratio & 41,930 & 32,145 & 10,430 & 24.9 \\
High sodium & 57,390 & 43,241 & 3,900 & 6.8 \\
Low sodium & 54,759 & 41,526 & 8,457 & 15.4 \\
High potassium & 55,915 & 42,348 & 6,168 & 11.0 \\
Low potassium & 56,570 & 42,678 & 8,118 & 14.4 \\
High glucose & 39,084 & 30,204 & 14,428 & 36.9 \\
Low glucose & 57,967 & 43,715 & 3,732 & 6.4 \\
High creatinine & 48,973 & 37,556 & 6,815 & 13.9 \\
High urea & 46,286 & 35,609 & 9,241 & 20.0 \\
Low albumin & 53,539 & 40,602 & 7,748 & 14.5 \\
High alanine transaminase & 54,017 & 40,906 & 4,752 & 8.8 \\
High aspartate aminotransferase & 53,125 & 40,161 & 5,246 & 9.9 \\
High lactate dehydrogenase & 55,094 & 41,749 & 4,974 & 9.0 \\
High bilirubin & 56,066 & 42,212 & 2,611 & 4.7 \\
High cholesterol & 58,307 & 43,895 & 475 & 0.8 \\
High triglyceride & 58,083 & 43,736 & 1,632 & 2.8 \\
High ferritin & 57,909 & 43,615 & 2,267 & 3.9 \\
High creatinine kinase & 57,037 & 42,883 & 1,508 & 2.6 \\
High C-reactive protein & 57,911 & 43,667 & 1,728 & 3.0 \\
\textbf{Clinical outcomes} & & & & \\
Long length of stay ($\geq$ 7 days) & 58,513 & 44,055 & 17,218 & 29.4 \\
Readmission within 30 days & 58,512 & 44,055 & 3,143 & 5.4 \\
Mortality & 58,513 & 44,055 & 1,741 & 3.0 \\
\end{longtable}\endgroup

Abbreviations: MIMIC, Medical Information Mart for Intensive Care; SickKids, The Hospital for Sick Children.

\clearpage
\subsection*{Supplementary Table S3. OMOP-derived event description templates and source columns}

\begingroup\scriptsize\setlength{\tabcolsep}{4pt}\begin{longtable}[]{@{}
  >{\raggedright\arraybackslash}p{(\columnwidth - 4\tabcolsep) * \real{0.1839}}
  >{\raggedright\arraybackslash}p{(\columnwidth - 4\tabcolsep) * \real{0.2759}}
  >{\raggedright\arraybackslash}p{(\columnwidth - 4\tabcolsep) * \real{0.5172}}@{}}
\toprule\noalign{}
\begin{minipage}[b]{\linewidth}\raggedright
\textbf{Category}
\end{minipage} & \begin{minipage}[b]{\linewidth}\centering
\textbf{Template segments}
\end{minipage} & \begin{minipage}[b]{\linewidth}\centering
\textbf{Slot mapping to OMOP-derived fields\textsuperscript{b}}
\end{minipage} \\
\midrule\noalign{}
\endhead
\bottomrule\noalign{}
\endlastfoot
demographics & Patient Birth Date\textsuperscript{a} & None \\
demographics & Patient Death Date\textsuperscript{a} & None \\
demographics & Patient Sex: \{sex\} & sex: gender\_concept\_name \\
Measurement* & \begin{minipage}[t]{\linewidth}\raggedright
Measurement: \{name\}\\
Result: \{result\}\strut
\end{minipage} & \begin{minipage}[t]{\linewidth}\raggedright
name: measurement\_concept\_name\\
result: meas\_value\_concept\_name\strut
\end{minipage} \\
Condition & Condition: \{name\} & name: condition\_concept\_name \\
Observation & Observation: \{name\} & name: observation\_concept\_name \\
Procedure & Procedure: \{name\} & name: procedure\_concept\_name \\
Drug* & \begin{minipage}[t]{\linewidth}\raggedright
Drug: \{name\}\\
Route: \{route\}\strut
\end{minipage} & \begin{minipage}[t]{\linewidth}\raggedright
name: drug\_concept\_name\\
route: drug\_route\_concept\_name\strut
\end{minipage} \\
note & Note: \{name\} & name: note\_concept\_name \\
specimen & Specimen: \{name\} & name: spec\_anatomic\_site\_concept\_name \\
visit & Visit: \{name\} & name: visit\_concept\_name \\
\end{longtable}\endgroup

*Template segments whose source columns are NULL are dropped. For example, measurement events emit "Measurement: \{name\}. Result: \{result\}" when meas\_value\_concept\_name is not null, and only "Measurement: \{name\}" when meas\_value\_concept\_name is null.

\textsuperscript{a}Patient birth date and patient death date were represented as fixed demographic event descriptions.

\textsuperscript{b}Fields ending in \_concept\_name are derived fields obtained by joining the corresponding OMOP concept identifier field, such as measurement\_concept\_id, condition\_concept\_id, or drug\_concept\_id, to the OMOP concept table and using the associated concept\_name. Analogous SEDAR-derived concept and attribute fields were used to construct event descriptions for cross-vocabulary evaluation.

Abbreviations: OMOP, Observational Medical Outcomes Partnership; SEDAR, SickKids Enterprise-wide Data in Azure Repository.

\clearpage
\subsection*{Supplementary Figure S2. Pretraining loss curves for PORTER using different text encoders}

\includegraphics[width=\linewidth,height=0.85\textheight,keepaspectratio]{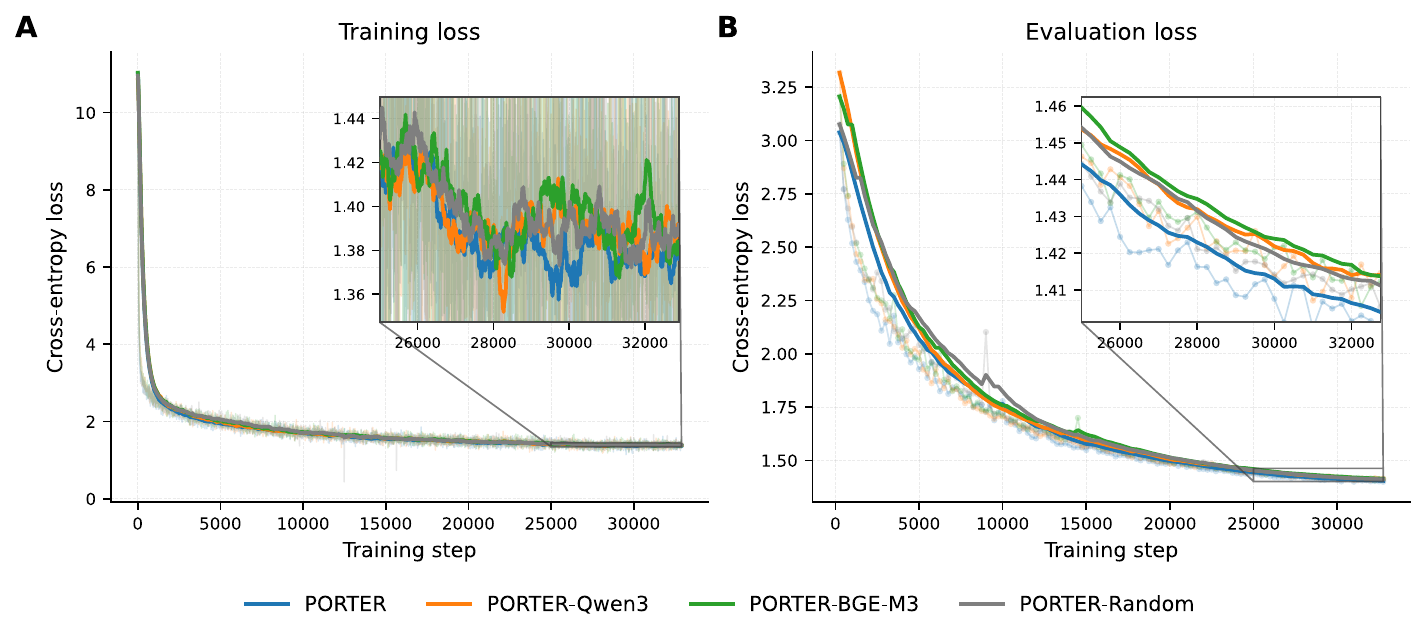}

Cross-entropy loss during pretraining for the four text encoder variants. Training loss was logged every 10 steps. Held-out evaluation loss was logged every 250 steps. Faded traces show raw values. Bold traces show exponential moving averages (span 50 for training, 10 for evaluation). Upper-right insets magnify the last \textasciitilde7,000 steps so late-training differences are legible.

\clearpage
\subsection*{Supplementary Figure S3. Empirical distribution of pre-clip numeric scalar features}

\includegraphics[width=\linewidth,height=0.85\textheight,keepaspectratio]{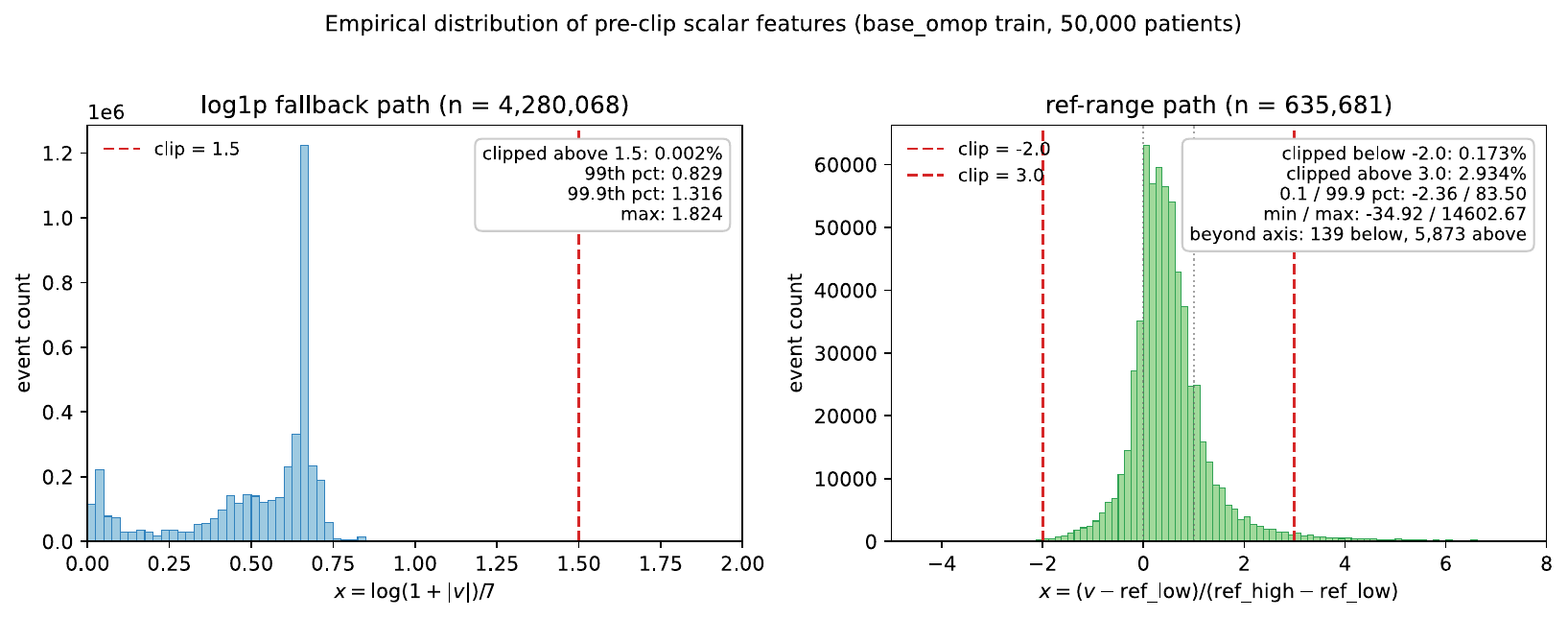}

Distributions of the log1p fallback scalar (left) and reference-range position scalar (right) are shown before clipping in a sample (n=50,000) from the SickKids training split.

\clearpage
\subsection*{Supplementary Table S4. Training, model, and evaluation hyperparameters}

\begingroup\scriptsize\setlength{\tabcolsep}{4pt}\begin{longtable}[]{@{}
  >{\raggedright\arraybackslash}p{(\columnwidth - 2\tabcolsep) * \real{0.5946}}
  >{\raggedright\arraybackslash}p{(\columnwidth - 2\tabcolsep) * \real{0.4054}}@{}}
\toprule\noalign{}
\begin{minipage}[b]{\linewidth}\raggedright
\textbf{Hyperparameter}
\end{minipage} & \begin{minipage}[b]{\linewidth}\centering
\textbf{Value}
\end{minipage} \\
\midrule\noalign{}
\endhead
\bottomrule\noalign{}
\endlastfoot
\textbf{Pretraining optimization and batching} & \\
\begin{minipage}[t]{\linewidth}\raggedright
\begin{quote}
Optimizer
\end{quote}
\end{minipage} & AdamW \\
\begin{minipage}[t]{\linewidth}\raggedright
\begin{quote}
Learning rate
\end{quote}
\end{minipage} & 5e-4 \\
\begin{minipage}[t]{\linewidth}\raggedright
\begin{quote}
Learning rate scheduler
\end{quote}
\end{minipage} & cosine\_with\_min\_lr \\
\begin{minipage}[t]{\linewidth}\raggedright
\begin{quote}
Num epochs
\end{quote}
\end{minipage} & 5 \\
\begin{minipage}[t]{\linewidth}\raggedright
\begin{quote}
Gradient accumulation steps
\end{quote}
\end{minipage} & 2 \\
\begin{minipage}[t]{\linewidth}\raggedright
\begin{quote}
Early stopping
\end{quote}
\end{minipage} & None \\
\begin{minipage}[t]{\linewidth}\raggedright
\begin{quote}
*Max clinical events per batch
\end{quote}
\end{minipage} & 16,384 \\
\begin{minipage}[t]{\linewidth}\raggedright
\begin{quote}
*Min patients per batch
\end{quote}
\end{minipage} & 1 \\
\begin{minipage}[t]{\linewidth}\raggedright
\begin{quote}
Weight decay
\end{quote}
\end{minipage} & 0.05 \\
\begin{minipage}[t]{\linewidth}\raggedright
\begin{quote}
Max gradient norm (clipping)
\end{quote}
\end{minipage} & 1.0 \\
\begin{minipage}[t]{\linewidth}\raggedright
\begin{quote}
Warmup steps
\end{quote}
\end{minipage} & 150 \\
\begin{minipage}[t]{\linewidth}\raggedright
\begin{quote}
Adam $\beta$1
\end{quote}
\end{minipage} & 0.9 \\
\begin{minipage}[t]{\linewidth}\raggedright
\begin{quote}
Adam $\beta$2
\end{quote}
\end{minipage} & 0.95 \\
\begin{minipage}[t]{\linewidth}\raggedright
\begin{quote}
Floating-point format
\end{quote}
\end{minipage} & bf16 \\
\textbf{Transformer backbone} & \\
\begin{minipage}[t]{\linewidth}\raggedright
\begin{quote}
Hidden size
\end{quote}
\end{minipage} & 768 \\
\begin{minipage}[t]{\linewidth}\raggedright
\begin{quote}
Num layers
\end{quote}
\end{minipage} & 28 \\
\begin{minipage}[t]{\linewidth}\raggedright
\begin{quote}
Num attention heads
\end{quote}
\end{minipage} & 12 \\
\begin{minipage}[t]{\linewidth}\raggedright
\begin{quote}
Intermediate size
\end{quote}
\end{minipage} & 768 \\
\begin{minipage}[t]{\linewidth}\raggedright
\begin{quote}
Activation
\end{quote}
\end{minipage} & SwiGLU \\
\begin{minipage}[t]{\linewidth}\raggedright
\begin{quote}
**Alternating dense layers
\end{quote}
\end{minipage} & Yes \\
\begin{minipage}[t]{\linewidth}\raggedright
\begin{quote}
**Attention width
\end{quote}
\end{minipage} & 128 \\
\textbf{Linear probe (Logistic Regression)} & \\
\begin{minipage}[t]{\linewidth}\raggedright
\begin{quote}
Input preprocessing
\end{quote}
\end{minipage} & StandardScaler \\
\begin{minipage}[t]{\linewidth}\raggedright
\begin{quote}
Solver
\end{quote}
\end{minipage} & LBFGS \\
\begin{minipage}[t]{\linewidth}\raggedright
\begin{quote}
Regularization
\end{quote}
\end{minipage} & L2 \\
\begin{minipage}[t]{\linewidth}\raggedright
\begin{quote}
Inverse regularization (C)
\end{quote}
\end{minipage} & 1, 0.1, 0.01, 0.001, 0.0001 \\
\begin{minipage}[t]{\linewidth}\raggedright
\begin{quote}
Max iterations
\end{quote}
\end{minipage} & 10,000 \\
\end{longtable}\endgroup

\textbf{*} Batches were constructed using a fixed clinical-event budget (16,384 clinical events per batch) with a minimum of one patient per batch to accommodate variable-length patient sequences without padding. This imposes an effective upper bound of 16,384 events per patient sequence (i.e., maximum context window). When multiple patient sequences were in a batch, causal masking prevented attention across patient boundaries, ensuring independent sequence modeling.

** The transformer alternates between global and local self-attention, starting with a global attention layer, followed by a local attention layer (attention width of 128), and repeating this pattern throughout.

\clearpage
\subsection*{Supplementary Figure S4. Illustrative serialized patient timeline for the text serialization comparator}

\includegraphics[width=\linewidth,height=0.85\textheight,keepaspectratio]{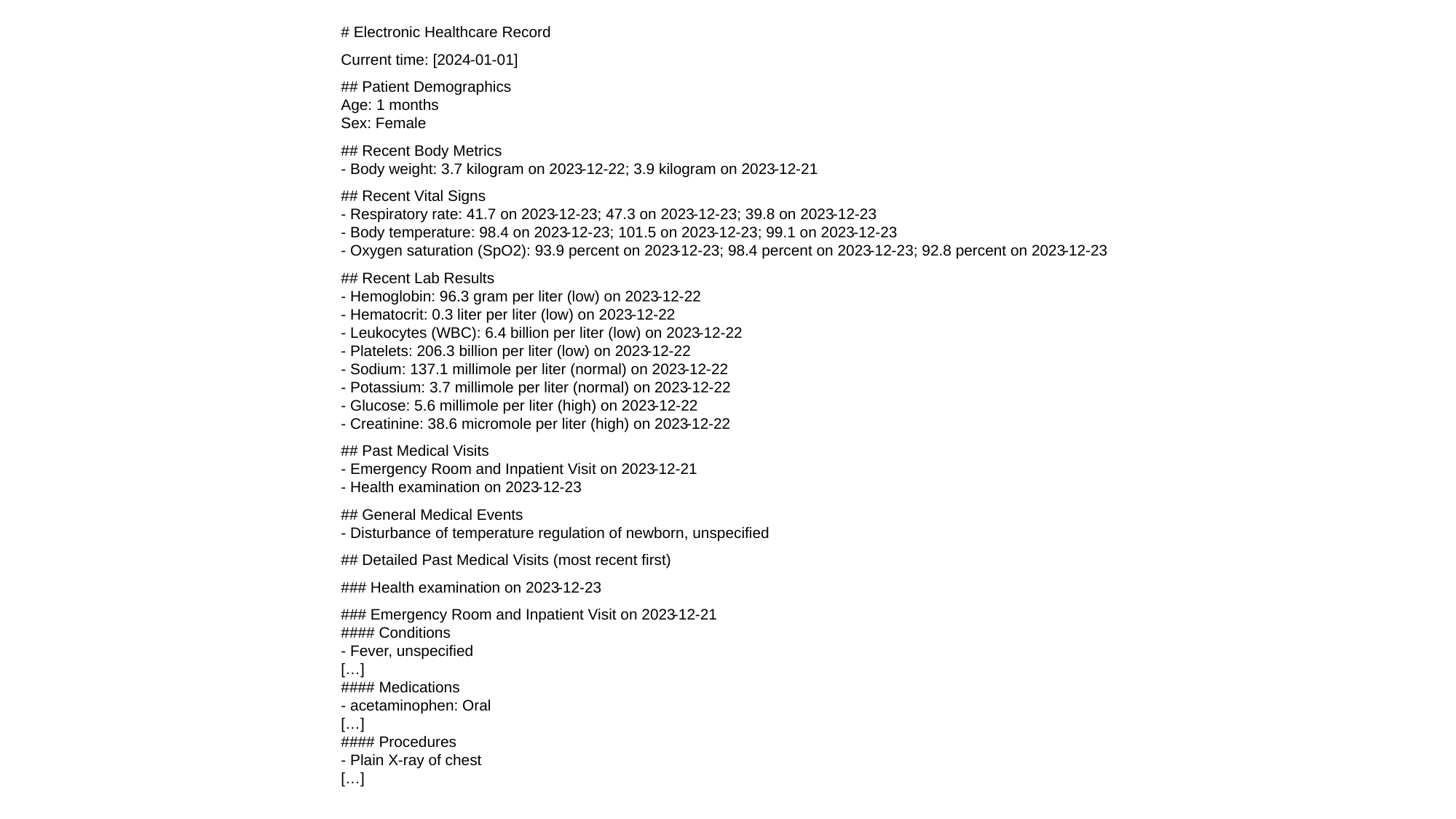}

Text serialization of a single synthetic patient\textquotesingle s event stream into Markdown for the text serialization comparator. The full document is prefixed with the task-specific instruction (Supplementary Table S4) before encoding with Qwen3-Embedding-8B.

Abbreviation: LLM, large language model.

\clearpage
\subsection*{Supplementary Table S5. Task-specific retrieval instructions used by the text serialization comparator}

\begingroup\scriptsize\setlength{\tabcolsep}{4pt}\begin{longtable}[]{@{}
  >{\raggedright\arraybackslash}p{(\columnwidth - 2\tabcolsep) * \real{0.3654}}
  >{\raggedright\arraybackslash}p{(\columnwidth - 2\tabcolsep) * \real{0.6154}}@{}}
\toprule\noalign{}
\begin{minipage}[b]{\linewidth}\raggedright
\textbf{Task}
\end{minipage} & \begin{minipage}[b]{\linewidth}\centering
\textbf{Query (verbatim)}
\end{minipage} \\
\midrule\noalign{}
\endhead
\bottomrule\noalign{}
\endlastfoot
\textbf{Transfusions} & \\
Platelet transfusion & will the patient need a platelet transfusion \\
Red cell transfusion & will the patient need a red blood cell transfusion \\
\textbf{Procedure} & \\
Invasive intubation & will the patient require invasive intubation \\
Gastrostomy tube & will the patient need a gastrostomy tube placement \\
Echocardiogram & will the patient need an echocardiogram \\
Pulmonary function test & will the patient need a pulmonary function test \\
Lumbar puncture & will the patient need a lumbar puncture \\
Surgery & will the patient need surgery \\
Interventional radiology & will the patient need an image-guided therapy procedure \\
\textbf{Imaging} & \\
Plain radiography chest & will the patient need a chest x-ray \\
Ultrasound abdomen & will the patient need an abdominal ultrasound \\
Computerized tomography chest & will the patient need a chest CT scan \\
Computerized tomography abdomen & will the patient need an abdominal CT scan \\
Computerized tomography head & will the patient need a head CT scan \\
MRI head & will the patient need a head MRI \\
MRI whole body & will the patient need a whole body MRI \\
PET & will the patient need a PET scan \\
\textbf{Laboratory abnormality} & \\
High white blood count & will the patient have a high white blood cell count \\
Low white blood count & will the patient have a low white blood cell count \\
High absolute neutrophil count & will the patient have a high absolute neutrophil count \\
Low absolute neutrophil count & will the patient have a low absolute neutrophil count \\
High bands & will the patient have a high band neutrophil count \\
High lymphocyte & will the patient have a high lymphocyte count \\
Low lymphocyte & will the patient have a low lymphocyte count \\
High hemoglobin & will the patient have a high hemoglobin level \\
Low hemoglobin & will the patient have a low hemoglobin level \\
High mean corpuscular volume & will the patient have a high mean corpuscular volume \\
Low mean corpuscular volume & will the patient have a low mean corpuscular volume \\
High reticulocyte count & will the patient have a high absolute reticulocyte count \\
Low reticulocyte count & will the patient have a low absolute reticulocyte count \\
High platelet & will the patient have a high platelet count \\
Low platelet & will the patient have a low platelet count \\
High immature platelet fraction & will the patient have a high immature platelet fraction \\
Low immature platelet fraction & will the patient have a low immature platelet fraction \\
High mean platelet volume & will the patient have a high mean platelet volume \\
Low mean platelet volume & will the patient have a low mean platelet volume \\
High fibrinogen & will the patient have a high fibrinogen level \\
Low fibrinogen & will the patient have a low fibrinogen level \\
High partial thromboplastin time & will the patient have a high partial thromboplastin time \\
High international normalized ratio & will the patient have a high international normalized ratio \\
High sodium & will the patient have a high sodium level \\
Low sodium & will the patient have a low sodium level \\
High potassium & will the patient have a high potassium level \\
Low potassium & will the patient have a low potassium level \\
High glucose & will the patient have a high blood glucose level \\
Low glucose & will the patient have a low blood glucose level \\
High creatinine & will the patient have a high creatinine level \\
High urea & will the patient have a high urea level \\
Low albumin & will the patient have a low albumin level \\
High alanine transaminase & will the patient have a high alanine aminotransferase level \\
High aspartate aminotransferase & will the patient have a high aspartate aminotransferase level \\
High lactate dehydrogenase & will the patient have a high lactate dehydrogenase level \\
High bilirubin & will the patient have a high bilirubin level \\
High cholesterol & will the patient have a high cholesterol level \\
High triglyceride & will the patient have a high triglyceride level \\
High ferritin & will the patient have a high ferritin level \\
High creatinine kinase & will the patient have a high creatine kinase level \\
High C-reactive protein & will the patient have a high C-reactive protein level \\
High erythrocyte sedimentation rate & will the patient have a high erythrocyte sedimentation rate \\
Low PaO2 & will the patient have a low partial pressure of oxygen \\
Low SpO2 & will the patient have low oxygen saturation \\
\textbf{Medications} & \\
Any antibacterial & will the patient need antibacterial medication \\
Any antifungal & will the patient need antifungal medication \\
Any chemotherapy & will the patient need chemotherapy \\
Any antiepileptics & will the patient need antiepileptic medication \\
Any glucocorticoid & will the patient need glucocorticoid medication \\
Dexamethasone & will the patient need dexamethasone \\
Any opioid & will the patient need opioid medication \\
Morphine & will the patient need morphine \\
Fentanyl & will the patient need fentanyl \\
Any inotrope & will the patient need inotrope or vasopressor medication \\
\textbf{Clinical outcomes} & \\
Long length of stay ($\geq$ 7 days) & will the patient stay in the hospital for more than 7 days \\
Readmission within 30 days & will the patient be readmitted to the hospital within 30 days \\
Mortality & will the patient die during this hospital admission \\
\end{longtable}\endgroup

Abbreviations: LLM, large language model.

\textbf{\hfill\break
}

\clearpage
\subsection*{Supplementary Table S6. Pretraining compute and parameter size by model}

\begingroup\scriptsize\setlength{\tabcolsep}{4pt}\begin{longtable}[]{@{}
  >{\raggedright\arraybackslash}p{(\columnwidth - 8\tabcolsep) * \real{0.1600}}
  >{\raggedright\arraybackslash}p{(\columnwidth - 8\tabcolsep) * \real{0.2160}}
  >{\raggedright\arraybackslash}p{(\columnwidth - 8\tabcolsep) * \real{0.2320}}
  >{\raggedright\arraybackslash}p{(\columnwidth - 8\tabcolsep) * \real{0.2320}}
  >{\raggedright\arraybackslash}p{(\columnwidth - 8\tabcolsep) * \real{0.1440}}@{}}
\toprule\noalign{}
\begin{minipage}[b]{\linewidth}\raggedright
\textbf{Model*}
\end{minipage} & \begin{minipage}[b]{\linewidth}\centering
\textbf{Frozen lookup params}
\end{minipage} & \begin{minipage}[b]{\linewidth}\centering
\textbf{Trainable input params}
\end{minipage} & \begin{minipage}[b]{\linewidth}\centering
\textbf{Total trainable params}
\end{minipage} & \begin{minipage}[b]{\linewidth}\centering
\textbf{Total FLOPs}
\end{minipage} \\
\midrule\noalign{}
\endhead
\bottomrule\noalign{}
\endlastfoot
PORTER & 15.12 M & 1.64 M & 192.87 M & 1.01 $\times$ 10\textsuperscript{18} \\
Fixed-Vocab FM & 0 & 75.50 M & 266.72 M & 9.98 $\times$ 10\textsuperscript{17} \\
PORTER-Qwen3 & 80.63 M & 4.20 M & 195.42 M & 1.02 $\times$ 10\textsuperscript{18} \\
PORTER-BGE-M3 & 20.16 M & 1.84 M & 193.06 M & 1.01 $\times$ 10\textsuperscript{18} \\
PORTER-Random & 80.63 M & 4.20 M & 195.42 M & 1.02 $\times$ 10\textsuperscript{18} \\
PORTER-NoNum & 15.12 M & 1.18 M & 192.41 M & 1.00 $\times$ 10\textsuperscript{18} \\
PORTER-NumText & 876.74 M & 1.18 M & 192.41 M & 1.00 $\times$ 10\textsuperscript{18} \\
\end{longtable}\endgroup

* The transformer backbone (115.63 M), next-event prediction head (75.60 M), and number of events seen (894 M) are fixed across all models. All parameter counts are reported in millions (M). Differences in trainable parameters arise from the input pathway: 1) the learnable joint embedding table (i.e., Fixed-Vocab FM, 75.50 M); 2) the projection MLP from each text encoder\textquotesingle s native dimension to the backbone hidden dimension; 3) the FiLM numeric pathway present in all PORTER variants except PORTER-NoNum and PORTER-NumText, which render numeric values as text or omit them entirely.

Abbreviations: FiLM, feature-wise linear modulation; FLOPs, floating-point operations; FM, foundation model; M, million; MLP, multilayer perceptron.

\begin{landscape}
\subsection*{Supplementary Table S7. Per-task AUROC for PORTER and Fixed-Vocab FM across in-domain, cross-vocabulary, and cross-site settings}

\begingroup\scriptsize\setlength{\tabcolsep}{4pt}\begin{longtable}[]{@{}
  >{\raggedright\arraybackslash}p{(\columnwidth - 16\tabcolsep) * \real{0.1557}}
  >{\centering\arraybackslash}p{(\columnwidth - 16\tabcolsep) * \real{0.0861}}
  >{\centering\arraybackslash}p{(\columnwidth - 16\tabcolsep) * \real{0.0861}}
  >{\centering\arraybackslash}p{(\columnwidth - 16\tabcolsep) * \real{0.1148}}
  >{\centering\arraybackslash}p{(\columnwidth - 16\tabcolsep) * \real{0.0984}}
  >{\centering\arraybackslash}p{(\columnwidth - 16\tabcolsep) * \real{0.1311}}
  >{\centering\arraybackslash}p{(\columnwidth - 16\tabcolsep) * \real{0.1025}}
  >{\centering\arraybackslash}p{(\columnwidth - 16\tabcolsep) * \real{0.0861}}
  >{\centering\arraybackslash}p{(\columnwidth - 16\tabcolsep) * \real{0.1311}}@{}}
\toprule\noalign{}
\begin{minipage}[b]{\linewidth}\raggedright
\textbf{Task}
\end{minipage} & \multicolumn{2}{c}{\textbf{In-domain (SK)}} & \multicolumn{3}{c}{\textbf{Cross-vocabulary (SK)}} & \multicolumn{3}{c}{\textbf{Cross-site (MIMIC)}} \\
 & \textbf{PORTER} & \textbf{Fixed-Vocab FM} & \textbf{PORTER} & \textbf{Fixed-Vocab FM\textsuperscript{a}} & \begin{minipage}[t]{\linewidth}\centering
\textbf{Target-vocabulary\\
Fixed-Vocab FM reference\textsuperscript{b}}\strut
\end{minipage} & \textbf{PORTER} & \textbf{Fixed-Vocab FM} & \begin{minipage}[t]{\linewidth}\centering
\textbf{Target-site\\
Fixed-Vocab FM reference\textsuperscript{c}}\strut
\end{minipage} \\
\midrule\noalign{}
\endhead
\bottomrule\noalign{}
\endlastfoot
\textbf{Transfusions} & & & & & & & & \\
Platelet transfusion & 0.967 & 0.970 & 0.944 & -\/- & 0.961 & -\/- & -\/- & -\/- \\
Red cell transfusion & 0.929 & 0.926 & 0.891 & -\/- & 0.919 & -\/- & -\/- & -\/- \\
\textbf{Procedure} & & & & & & & & \\
Invasive intubation & 0.924 & 0.925 & 0.894 & -\/- & 0.931 & -\/- & -\/- & -\/- \\
Gastrostomy tube & 0.885 & 0.886 & 0.855 & -\/- & 0.903 & -\/- & -\/- & -\/- \\
Echocardiogram & 0.894 & 0.892 & 0.869 & -\/- & 0.890 & -\/- & -\/- & -\/- \\
Pulmonary function test & 0.966 & 0.966 & 0.941 & -\/- & 0.962 & -\/- & -\/- & -\/- \\
Lumbar puncture & 0.947 & 0.941 & 0.916 & -\/- & 0.942 & -\/- & -\/- & -\/- \\
Surgery & 0.891 & 0.887 & 0.850 & -\/- & 0.894 & -\/- & -\/- & -\/- \\
Interventional radiology & 0.864 & 0.864 & 0.815 & -\/- & 0.860 & -\/- & -\/- & -\/- \\
\textbf{Imaging} & & & & & & & & \\
Plain radiography chest & 0.820 & 0.820 & 0.787 & -\/- & 0.824 & -\/- & -\/- & -\/- \\
Ultrasound abdomen & 0.841 & 0.847 & 0.810 & -\/- & 0.836 & -\/- & -\/- & -\/- \\
Computerized tomography chest & 0.868 & 0.866 & 0.818 & -\/- & 0.852 & -\/- & -\/- & -\/- \\
Computerized tomography abdomen & 0.894 & 0.891 & 0.824 & -\/- & 0.883 & -\/- & -\/- & -\/- \\
Computerized tomography head & 0.910 & 0.909 & 0.869 & -\/- & 0.911 & -\/- & -\/- & -\/- \\
MRI head & 0.908 & 0.912 & 0.875 & -\/- & 0.912 & -\/- & -\/- & -\/- \\
MRI whole body & 0.948 & 0.944 & 0.903 & -\/- & 0.913 & -\/- & -\/- & -\/- \\
PET & 0.897 & 0.901 & 0.823 & -\/- & 0.801 & -\/- & -\/- & -\/- \\
\textbf{Laboratory abnormality} & & & & & & & & \\
High white blood count & 0.834 & 0.830 & 0.781 & -\/- & 0.795 & 0.760 & 0.751 & 0.803 \\
Low white blood count & 0.920 & 0.916 & 0.883 & -\/- & 0.895 & 0.869 & 0.876 & 0.905 \\
High absolute neutrophil count & 0.839 & 0.840 & 0.788 & -\/- & 0.814 & 0.795 & 0.761 & 0.818 \\
Low absolute neutrophil count & 0.934 & 0.934 & 0.906 & -\/- & 0.917 & 0.853 & 0.852 & 0.886 \\
High bands & 0.872 & 0.874 & 0.845 & -\/- & 0.868 & -\/- & -\/- & -\/- \\
High lymphocyte & 0.852 & 0.848 & 0.816 & -\/- & 0.830 & 0.713 & 0.680 & 0.737 \\
Low lymphocyte & 0.894 & 0.895 & 0.871 & -\/- & 0.887 & 0.816 & 0.793 & 0.831 \\
High hemoglobin & 0.861 & 0.848 & 0.818 & -\/- & 0.839 & -\/- & -\/- & -\/- \\
Low hemoglobin & 0.887 & 0.882 & 0.855 & -\/- & 0.875 & 0.859 & 0.850 & 0.876 \\
High mean corpuscular volume & 0.892 & 0.895 & 0.860 & -\/- & 0.885 & 0.793 & 0.839 & 0.813 \\
Low mean corpuscular volume & 0.856 & 0.838 & 0.779 & -\/- & 0.804 & 0.806 & 0.856 & 0.766 \\
High reticulocyte count & 0.898 & 0.894 & 0.855 & -\/- & 0.884 & -\/- & -\/- & -\/- \\
Low reticulocyte count & 0.893 & 0.887 & 0.851 & -\/- & 0.851 & -\/- & -\/- & -\/- \\
High platelet & 0.841 & 0.839 & 0.805 & -\/- & 0.827 & 0.792 & 0.773 & 0.835 \\
Low platelet & 0.877 & 0.880 & 0.846 & -\/- & 0.875 & 0.813 & 0.800 & 0.852 \\
High immature platelet fraction & 0.892 & 0.885 & 0.855 & -\/- & 0.882 & -\/- & -\/- & -\/- \\
Low immature platelet fraction & 0.870 & 0.873 & 0.844 & -\/- & 0.851 & -\/- & -\/- & -\/- \\
High mean platelet volume & 0.902 & 0.902 & 0.880 & -\/- & 0.907 & -\/- & -\/- & -\/- \\
Low mean platelet volume & 0.792 & 0.797 & 0.762 & -\/- & 0.783 & -\/- & -\/- & -\/- \\
High fibrinogen & 0.891 & 0.888 & 0.848 & -\/- & 0.884 & 0.753 & 0.713 & 0.794 \\
Low fibrinogen & 0.920 & 0.909 & 0.886 & -\/- & 0.899 & 0.866 & 0.843 & 0.895 \\
High partial thromboplastin time & 0.899 & 0.901 & 0.865 & -\/- & 0.895 & 0.867 & 0.844 & 0.893 \\
High international normalized ratio & 0.869 & 0.874 & 0.834 & -\/- & 0.863 & 0.884 & 0.869 & 0.906 \\
High sodium & 0.851 & 0.849 & 0.813 & -\/- & 0.840 & 0.818 & 0.812 & 0.836 \\
Low sodium & 0.854 & 0.854 & 0.832 & -\/- & 0.841 & 0.825 & 0.811 & 0.834 \\
High potassium & 0.836 & 0.834 & 0.807 & -\/- & 0.833 & 0.834 & 0.815 & 0.845 \\
Low potassium & 0.868 & 0.871 & 0.838 & -\/- & 0.865 & 0.819 & 0.823 & 0.835 \\
High glucose & 0.832 & 0.834 & 0.796 & -\/- & 0.830 & 0.927 & 0.920 & 0.935 \\
Low glucose & 0.882 & 0.884 & 0.855 & -\/- & 0.873 & 0.783 & 0.761 & 0.812 \\
High creatinine & 0.914 & 0.910 & 0.883 & -\/- & 0.896 & 0.862 & 0.822 & 0.868 \\
High urea & 0.926 & 0.928 & 0.896 & -\/- & 0.909 & 0.877 & 0.864 & 0.887 \\
Low albumin & 0.878 & 0.877 & 0.849 & -\/- & 0.869 & 0.849 & 0.824 & 0.872 \\
High alanine transaminase & 0.885 & 0.886 & 0.853 & -\/- & 0.865 & 0.800 & 0.810 & 0.826 \\
High aspartate aminotransferase & 0.894 & 0.896 & 0.862 & -\/- & 0.878 & 0.800 & 0.784 & 0.829 \\
High lactate dehydrogenase & 0.913 & 0.910 & 0.875 & -\/- & 0.897 & 0.827 & 0.810 & 0.854 \\
High bilirubin & 0.893 & 0.885 & 0.862 & -\/- & 0.879 & 0.834 & 0.811 & 0.862 \\
High cholesterol & 0.906 & 0.908 & 0.864 & -\/- & 0.882 & 0.785 & 0.754 & 0.861 \\
High triglyceride & 0.876 & 0.858 & 0.817 & -\/- & 0.843 & 0.773 & 0.770 & 0.830 \\
High ferritin & 0.870 & 0.870 & 0.834 & -\/- & 0.856 & 0.833 & 0.801 & 0.852 \\
High creatinine kinase & 0.905 & 0.911 & 0.862 & -\/- & 0.893 & 0.809 & 0.802 & 0.853 \\
High C-reactive protein & 0.849 & 0.851 & 0.814 & -\/- & 0.840 & 0.793 & 0.763 & 0.829 \\
High erythrocyte sedimentation rate & 0.932 & 0.936 & 0.884 & -\/- & 0.903 & -\/- & -\/- & -\/- \\
Low PaO2 & 0.954 & 0.958 & 0.943 & -\/- & 0.958 & -\/- & -\/- & -\/- \\
Low SpO2 & 0.815 & 0.817 & 0.766 & -\/- & 0.823 & -\/- & -\/- & -\/- \\
\textbf{Medications} & & & & & & & & \\
Any antibacterial & 0.864 & 0.863 & 0.831 & -\/- & 0.854 & -\/- & -\/- & -\/- \\
Any antifungal & 0.942 & 0.949 & 0.925 & -\/- & 0.941 & -\/- & -\/- & -\/- \\
Any chemotherapy & 0.974 & 0.977 & 0.952 & -\/- & 0.969 & -\/- & -\/- & -\/- \\
Any antiepileptics & 0.870 & 0.866 & 0.832 & -\/- & 0.866 & -\/- & -\/- & -\/- \\
Any glucocorticoid & 0.842 & 0.842 & 0.806 & -\/- & 0.844 & -\/- & -\/- & -\/- \\
Dexamethasone & 0.842 & 0.847 & 0.802 & -\/- & 0.848 & -\/- & -\/- & -\/- \\
Any opioid & 0.853 & 0.855 & 0.814 & -\/- & 0.856 & -\/- & -\/- & -\/- \\
Morphine & 0.836 & 0.832 & 0.799 & -\/- & 0.837 & -\/- & -\/- & -\/- \\
Fentanyl & 0.843 & 0.847 & 0.807 & -\/- & 0.846 & -\/- & -\/- & -\/- \\
Any inotrope & 0.926 & 0.930 & 0.880 & -\/- & 0.930 & -\/- & -\/- & -\/- \\
\textbf{Clinical outcomes} & & & & & & & & \\
Long length of stay ($\geq$ 7 days) & 0.834 & 0.835 & 0.795 & -\/- & 0.835 & 0.811 & 0.789 & 0.835 \\
Readmission within 30 days & 0.793 & 0.794 & 0.775 & -\/- & 0.786 & 0.833 & 0.819 & 0.844 \\
Mortality & 0.941 & 0.942 & 0.905 & -\/- & 0.937 & 0.895 & 0.886 & 0.932 \\
\end{longtable}\endgroup

\textsuperscript{a}Fixed-Vocab FM was not evaluated in cross-vocabulary transfer because SEDAR composite event tokens had no learned input embeddings.

\textsuperscript{b}The target-vocabulary reference is a Fixed-Vocab FM trained directly on the SEDAR vocabulary.

\textsuperscript{c}The target-site reference is a Fixed-Vocab FM trained directly on MIMIC. MIMIC cross-site columns include 36 labels.

Abbreviations: AUROC, area under the receiver operating characteristic curve; FM, foundation model; MIMIC, Medical Information Mart for Intensive Care; SEDAR, SickKids Enterprise-wide Data in Azure Repository; SickKids, The Hospital for Sick Children; SK, SickKids.

\end{landscape}
\clearpage
\subsection*{Supplementary Figure S5. Sample efficiency curves across in-domain, cross-vocabulary, and cross-site settings for PORTER and Fixed-Vocab FM}

\includegraphics[width=\linewidth,height=0.85\textheight,keepaspectratio]{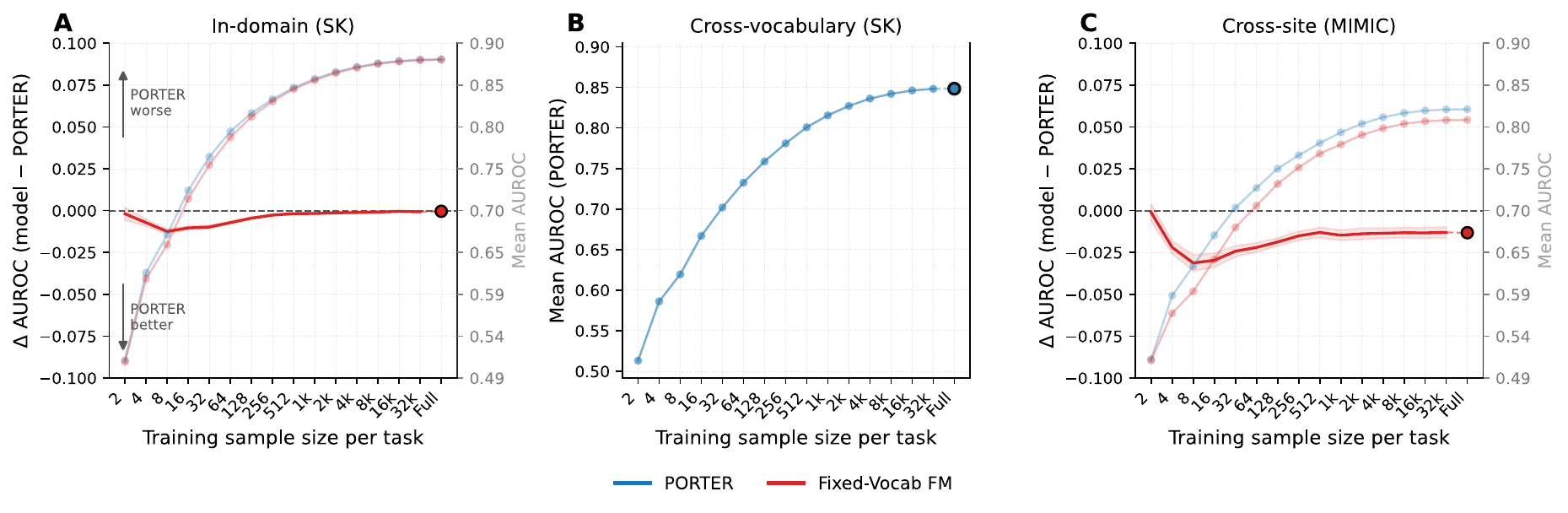}

Curves summarize linear-probe performance as labeled training examples per task increase. PORTER is compared with Fixed-Vocab FM when it can be evaluated. The cross-vocabulary panel shows PORTER alone because Fixed-Vocab FM was not evaluated in this setting. SEDAR composite event tokens had no learned input embeddings.

Abbreviations: FM, foundation model; MIMIC, Medical Information Mart for Intensive Care; SEDAR, SickKids Enterprise-wide Data in Azure Repository; SickKids, The Hospital for Sick Children; SK, SickKids.

\clearpage
\subsection*{Supplementary Table S8. Full-shot per-task AUROC for PORTER and text serialization comparator}

\begingroup\scriptsize\setlength{\tabcolsep}{4pt}\begin{longtable}[]{@{}
  >{\raggedright\arraybackslash}p{(\columnwidth - 4\tabcolsep) * \real{0.4086}}
  >{\raggedright\arraybackslash}p{(\columnwidth - 4\tabcolsep) * \real{0.1828}}
  >{\raggedright\arraybackslash}p{(\columnwidth - 4\tabcolsep) * \real{0.3871}}@{}}
\toprule\noalign{}
\begin{minipage}[b]{\linewidth}\raggedright
\textbf{Task}
\end{minipage} & \begin{minipage}[b]{\linewidth}\centering
\textbf{PORTER}
\end{minipage} & \begin{minipage}[b]{\linewidth}\centering
\textbf{Text serialization comparator}
\end{minipage} \\
\midrule\noalign{}
\endhead
\bottomrule\noalign{}
\endlastfoot
\textbf{Transfusions} & & \\
Platelet transfusion & 0.967 & 0.963 \\
Red cell transfusion & 0.929 & 0.915 \\
\textbf{Procedure} & & \\
Invasive intubation & 0.924 & 0.896 \\
Gastrostomy tube & 0.885 & 0.912 \\
Echocardiogram & 0.894 & 0.882 \\
Pulmonary function test & 0.966 & 0.948 \\
Lumbar puncture & 0.947 & 0.930 \\
Surgery & 0.891 & 0.872 \\
Interventional radiology & 0.864 & 0.841 \\
\textbf{Imaging} & & \\
Plain radiography chest & 0.820 & 0.797 \\
Ultrasound abdomen & 0.841 & 0.821 \\
Computerized tomography chest & 0.868 & 0.845 \\
Computerized tomography abdomen & 0.894 & 0.867 \\
Computerized tomography head & 0.910 & 0.893 \\
MRI head & 0.908 & 0.891 \\
MRI whole body & 0.948 & 0.911 \\
PET & 0.897 & 0.892 \\
\textbf{Laboratory abnormality} & & \\
High white blood count & 0.834 & 0.804 \\
Low white blood count & 0.920 & 0.903 \\
High absolute neutrophil count & 0.839 & 0.810 \\
Low absolute neutrophil count & 0.934 & 0.917 \\
High bands & 0.872 & 0.854 \\
High lymphocyte & 0.852 & 0.813 \\
Low lymphocyte & 0.894 & 0.874 \\
High hemoglobin & 0.861 & 0.848 \\
Low hemoglobin & 0.887 & 0.866 \\
High mean corpuscular volume & 0.892 & 0.895 \\
Low mean corpuscular volume & 0.856 & 0.836 \\
High reticulocyte count & 0.898 & 0.878 \\
Low reticulocyte count & 0.893 & 0.865 \\
High platelet & 0.841 & 0.833 \\
Low platelet & 0.877 & 0.877 \\
High immature platelet fraction & 0.892 & 0.886 \\
Low immature platelet fraction & 0.870 & 0.858 \\
High mean platelet volume & 0.902 & 0.906 \\
Low mean platelet volume & 0.792 & 0.781 \\
High fibrinogen & 0.891 & 0.858 \\
Low fibrinogen & 0.920 & 0.900 \\
High partial thromboplastin time & 0.899 & 0.882 \\
High international normalized ratio & 0.869 & 0.838 \\
High sodium & 0.851 & 0.814 \\
Low sodium & 0.854 & 0.834 \\
High potassium & 0.836 & 0.819 \\
Low potassium & 0.868 & 0.849 \\
High glucose & 0.832 & 0.804 \\
Low glucose & 0.882 & 0.849 \\
High creatinine & 0.914 & 0.905 \\
High urea & 0.926 & 0.916 \\
Low albumin & 0.878 & 0.852 \\
High alanine transaminase & 0.885 & 0.861 \\
High aspartate aminotransferase & 0.894 & 0.873 \\
High lactate dehydrogenase & 0.913 & 0.882 \\
High bilirubin & 0.893 & 0.877 \\
High cholesterol & 0.906 & 0.874 \\
High triglyceride & 0.876 & 0.817 \\
High ferritin & 0.870 & 0.846 \\
High creatinine kinase & 0.905 & 0.821 \\
High C-reactive protein & 0.849 & 0.823 \\
High erythrocyte sedimentation rate & 0.932 & 0.916 \\
Low PaO2 & 0.954 & 0.949 \\
Low SpO2 & 0.815 & 0.779 \\
\textbf{Medications} & & \\
Any antibacterial & 0.864 & 0.841 \\
Any antifungal & 0.942 & 0.935 \\
Any chemotherapy & 0.974 & 0.968 \\
Any antiepileptics & 0.870 & 0.844 \\
Any glucocorticoid & 0.842 & 0.806 \\
Dexamethasone & 0.842 & 0.809 \\
Any opioid & 0.853 & 0.821 \\
Morphine & 0.836 & 0.799 \\
Fentanyl & 0.843 & 0.809 \\
Any inotrope & 0.926 & 0.909 \\
\textbf{Clinical outcomes} & & \\
Long length of stay ($\geq$ 7 days) & 0.834 & 0.812 \\
Readmission within 30 days & 0.793 & 0.796 \\
Mortality & 0.941 & 0.925 \\
\end{longtable}\endgroup

Abbreviations: AUROC, area under the receiver operating characteristic curve; LLM, large language model..

\begin{landscape}
\subsection*{Supplementary Table S9. Per-task AUROC for PORTER and its text encoder ablations across in-domain, cross-vocabulary, and cross-site settings}

\begingroup\scriptsize\setlength{\tabcolsep}{4pt}\begin{longtable}[]{@{}
  >{\raggedright\arraybackslash}p{(\columnwidth - 24\tabcolsep) * \real{0.1652}}
  >{\centering\arraybackslash}p{(\columnwidth - 24\tabcolsep) * \real{0.0913}}
  >{\centering\arraybackslash}p{(\columnwidth - 24\tabcolsep) * \real{0.0522}}
  >{\centering\arraybackslash}p{(\columnwidth - 24\tabcolsep) * \real{0.0565}}
  >{\centering\arraybackslash}p{(\columnwidth - 24\tabcolsep) * \real{0.0565}}
  >{\centering\arraybackslash}p{(\columnwidth - 24\tabcolsep) * \real{0.1217}}
  >{\centering\arraybackslash}p{(\columnwidth - 24\tabcolsep) * \real{0.0522}}
  >{\centering\arraybackslash}p{(\columnwidth - 24\tabcolsep) * \real{0.0565}}
  >{\centering\arraybackslash}p{(\columnwidth - 24\tabcolsep) * \real{0.0565}}
  >{\centering\arraybackslash}p{(\columnwidth - 24\tabcolsep) * \real{0.1174}}
  >{\centering\arraybackslash}p{(\columnwidth - 24\tabcolsep) * \real{0.0522}}
  >{\centering\arraybackslash}p{(\columnwidth - 24\tabcolsep) * \real{0.0565}}
  >{\centering\arraybackslash}p{(\columnwidth - 24\tabcolsep) * \real{0.0565}}@{}}
\toprule\noalign{}
\begin{minipage}[b]{\linewidth}\raggedright
\textbf{Task}
\end{minipage} & \multicolumn{4}{c}{\textbf{In-domain (SK)}} & \multicolumn{4}{c}{\textbf{Cross-vocabulary (SK)}} & \multicolumn{4}{c}{\textbf{Cross-site (MIMIC)*}} \\
 & \textbf{BioLORD} & \textbf{Qwen3} & \textbf{BGE-M3} & \textbf{Random} & \textbf{BioLORD} & \textbf{Qwen3} & \textbf{BGE-M3} & \textbf{Random} & \textbf{BioLORD} & \textbf{Qwen3} & \textbf{BGE-M3} & \textbf{Random} \\
\midrule\noalign{}
\endhead
\bottomrule\noalign{}
\endlastfoot
\textbf{Transfusions} & & & & & & & & & & & & \\
Platelet transfusion & 0.967 & 0.968 & 0.969 & 0.964 & 0.944 & 0.939 & 0.930 & 0.902 & -\/- & -\/- & -\/- & -\/- \\
Red cell transfusion & 0.929 & 0.921 & 0.924 & 0.920 & 0.891 & 0.893 & 0.869 & 0.842 & -\/- & -\/- & -\/- & -\/- \\
\textbf{Procedure} & & & & & & & & & & & & \\
Invasive intubation & 0.924 & 0.925 & 0.908 & 0.920 & 0.894 & 0.884 & 0.848 & 0.813 & -\/- & -\/- & -\/- & -\/- \\
Gastrostomy tube & 0.885 & 0.884 & 0.884 & 0.867 & 0.855 & 0.842 & 0.784 & 0.765 & -\/- & -\/- & -\/- & -\/- \\
Echocardiogram & 0.894 & 0.889 & 0.882 & 0.889 & 0.869 & 0.864 & 0.844 & 0.800 & -\/- & -\/- & -\/- & -\/- \\
Pulmonary function test & 0.966 & 0.962 & 0.967 & 0.958 & 0.941 & 0.935 & 0.897 & 0.838 & -\/- & -\/- & -\/- & -\/- \\
Lumbar puncture & 0.947 & 0.946 & 0.936 & 0.938 & 0.916 & 0.898 & 0.866 & 0.816 & -\/- & -\/- & -\/- & -\/- \\
Surgery & 0.891 & 0.881 & 0.879 & 0.881 & 0.850 & 0.852 & 0.816 & 0.766 & -\/- & -\/- & -\/- & -\/- \\
Interventional radiology & 0.864 & 0.859 & 0.855 & 0.851 & 0.815 & 0.818 & 0.778 & 0.734 & -\/- & -\/- & -\/- & -\/- \\
\textbf{Imaging} & & & & & & & & & & & & \\
Plain radiography chest & 0.820 & 0.820 & 0.818 & 0.822 & 0.787 & 0.785 & 0.765 & 0.713 & -\/- & -\/- & -\/- & -\/- \\
Ultrasound abdomen & 0.841 & 0.836 & 0.836 & 0.839 & 0.810 & 0.802 & 0.774 & 0.736 & -\/- & -\/- & -\/- & -\/- \\
Computerized tomography chest & 0.868 & 0.860 & 0.855 & 0.849 & 0.818 & 0.808 & 0.801 & 0.753 & -\/- & -\/- & -\/- & -\/- \\
Computerized tomography abdomen & 0.894 & 0.891 & 0.891 & 0.883 & 0.824 & 0.847 & 0.793 & 0.738 & -\/- & -\/- & -\/- & -\/- \\
Computerized tomography head & 0.910 & 0.912 & 0.907 & 0.903 & 0.869 & 0.863 & 0.793 & 0.722 & -\/- & -\/- & -\/- & -\/- \\
MRI head & 0.908 & 0.905 & 0.906 & 0.905 & 0.875 & 0.855 & 0.812 & 0.739 & -\/- & -\/- & -\/- & -\/- \\
MRI whole body & 0.948 & 0.940 & 0.938 & 0.952 & 0.903 & 0.845 & 0.874 & 0.778 & -\/- & -\/- & -\/- & -\/- \\
PET & 0.897 & 0.917 & 0.916 & 0.880 & 0.823 & 0.829 & 0.795 & 0.750 & -\/- & -\/- & -\/- & -\/- \\
\textbf{Laboratory abnormality} & & & & & & & & & & & & \\
High white blood count & 0.834 & 0.833 & 0.831 & 0.835 & 0.781 & 0.775 & 0.749 & 0.697 & 0.760 & 0.765 & 0.756 & 0.758 \\
Low white blood count & 0.920 & 0.919 & 0.918 & 0.919 & 0.883 & 0.875 & 0.867 & 0.847 & 0.869 & 0.891 & 0.876 & 0.875 \\
High absolute neutrophil count & 0.839 & 0.838 & 0.836 & 0.838 & 0.788 & 0.785 & 0.754 & 0.721 & 0.795 & 0.785 & 0.776 & 0.758 \\
Low absolute neutrophil count & 0.934 & 0.934 & 0.933 & 0.931 & 0.906 & 0.902 & 0.884 & 0.864 & 0.853 & 0.860 & 0.850 & 0.856 \\
High bands & 0.872 & 0.870 & 0.871 & 0.869 & 0.845 & 0.838 & 0.822 & 0.809 & -\/- & -\/- & -\/- & -\/- \\
High lymphocyte & 0.852 & 0.848 & 0.849 & 0.849 & 0.816 & 0.817 & 0.798 & 0.761 & 0.713 & 0.716 & 0.689 & 0.704 \\
Low lymphocyte & 0.894 & 0.895 & 0.894 & 0.892 & 0.871 & 0.864 & 0.849 & 0.829 & 0.816 & 0.803 & 0.806 & 0.780 \\
High hemoglobin & 0.861 & 0.847 & 0.844 & 0.840 & 0.818 & 0.809 & 0.799 & 0.777 & -\/- & -\/- & -\/- & -\/- \\
Low hemoglobin & 0.887 & 0.886 & 0.882 & 0.881 & 0.855 & 0.856 & 0.832 & 0.805 & 0.859 & 0.862 & 0.865 & 0.855 \\
High mean corpuscular volume & 0.892 & 0.904 & 0.898 & 0.887 & 0.860 & 0.854 & 0.852 & 0.825 & 0.793 & 0.808 & 0.808 & 0.812 \\
Low mean corpuscular volume & 0.856 & 0.866 & 0.863 & 0.848 & 0.779 & 0.781 & 0.765 & 0.724 & 0.806 & 0.823 & 0.823 & 0.821 \\
High reticulocyte count & 0.898 & 0.890 & 0.891 & 0.894 & 0.855 & 0.855 & 0.846 & 0.824 & -\/- & -\/- & -\/- & -\/- \\
Low reticulocyte count & 0.893 & 0.876 & 0.886 & 0.872 & 0.851 & 0.839 & 0.821 & 0.793 & -\/- & -\/- & -\/- & -\/- \\
High platelet & 0.841 & 0.834 & 0.836 & 0.841 & 0.805 & 0.799 & 0.784 & 0.753 & 0.792 & 0.792 & 0.788 & 0.773 \\
Low platelet & 0.877 & 0.876 & 0.879 & 0.888 & 0.846 & 0.838 & 0.825 & 0.799 & 0.813 & 0.821 & 0.808 & 0.808 \\
High immature platelet fraction & 0.892 & 0.884 & 0.889 & 0.891 & 0.855 & 0.853 & 0.844 & 0.820 & -\/- & -\/- & -\/- & -\/- \\
Low immature platelet fraction & 0.870 & 0.861 & 0.868 & 0.859 & 0.844 & 0.827 & 0.825 & 0.794 & -\/- & -\/- & -\/- & -\/- \\
High mean platelet volume & 0.902 & 0.900 & 0.903 & 0.901 & 0.880 & 0.876 & 0.865 & 0.835 & -\/- & -\/- & -\/- & -\/- \\
Low mean platelet volume & 0.792 & 0.778 & 0.796 & 0.802 & 0.762 & 0.754 & 0.735 & 0.694 & -\/- & -\/- & -\/- & -\/- \\
High fibrinogen & 0.891 & 0.894 & 0.892 & 0.882 & 0.848 & 0.832 & 0.823 & 0.774 & 0.753 & 0.770 & 0.760 & 0.742 \\
Low fibrinogen & 0.920 & 0.913 & 0.904 & 0.907 & 0.886 & 0.878 & 0.840 & 0.823 & 0.866 & 0.869 & 0.875 & 0.867 \\
High partial thromboplastin time & 0.899 & 0.899 & 0.901 & 0.902 & 0.865 & 0.869 & 0.836 & 0.808 & 0.867 & 0.863 & 0.861 & 0.847 \\
High international normalized ratio & 0.869 & 0.871 & 0.868 & 0.870 & 0.834 & 0.826 & 0.799 & 0.769 & 0.884 & 0.882 & 0.883 & 0.873 \\
High sodium & 0.851 & 0.845 & 0.841 & 0.842 & 0.813 & 0.806 & 0.786 & 0.750 & 0.818 & 0.819 & 0.819 & 0.810 \\
Low sodium & 0.854 & 0.855 & 0.851 & 0.849 & 0.832 & 0.809 & 0.797 & 0.778 & 0.825 & 0.815 & 0.821 & 0.799 \\
High potassium & 0.836 & 0.829 & 0.833 & 0.829 & 0.807 & 0.797 & 0.785 & 0.764 & 0.834 & 0.828 & 0.828 & 0.816 \\
Low potassium & 0.868 & 0.868 & 0.863 & 0.864 & 0.838 & 0.838 & 0.818 & 0.778 & 0.819 & 0.811 & 0.808 & 0.805 \\
High glucose & 0.832 & 0.834 & 0.833 & 0.832 & 0.796 & 0.796 & 0.774 & 0.750 & 0.927 & 0.925 & 0.924 & 0.917 \\
Low glucose & 0.882 & 0.881 & 0.881 & 0.876 & 0.855 & 0.836 & 0.830 & 0.801 & 0.783 & 0.788 & 0.786 & 0.773 \\
High creatinine & 0.914 & 0.916 & 0.912 & 0.913 & 0.883 & 0.863 & 0.834 & 0.799 & 0.862 & 0.850 & 0.847 & 0.817 \\
High urea & 0.926 & 0.924 & 0.930 & 0.924 & 0.896 & 0.889 & 0.867 & 0.829 & 0.877 & 0.872 & 0.873 & 0.851 \\
Low albumin & 0.878 & 0.875 & 0.874 & 0.873 & 0.849 & 0.846 & 0.823 & 0.804 & 0.849 & 0.849 & 0.843 & 0.842 \\
High alanine transaminase & 0.885 & 0.885 & 0.886 & 0.882 & 0.853 & 0.831 & 0.808 & 0.789 & 0.800 & 0.810 & 0.803 & 0.794 \\
High aspartate aminotransferase & 0.894 & 0.899 & 0.894 & 0.894 & 0.862 & 0.836 & 0.816 & 0.793 & 0.800 & 0.807 & 0.797 & 0.784 \\
High lactate dehydrogenase & 0.913 & 0.910 & 0.906 & 0.905 & 0.875 & 0.855 & 0.843 & 0.841 & 0.827 & 0.829 & 0.830 & 0.821 \\
High bilirubin & 0.893 & 0.890 & 0.889 & 0.892 & 0.862 & 0.855 & 0.834 & 0.798 & 0.834 & 0.853 & 0.836 & 0.824 \\
High cholesterol & 0.906 & 0.890 & 0.899 & 0.907 & 0.864 & 0.823 & 0.833 & 0.702 & 0.785 & 0.782 & 0.790 & 0.789 \\
High triglyceride & 0.876 & 0.861 & 0.845 & 0.862 & 0.817 & 0.798 & 0.790 & 0.728 & 0.773 & 0.774 & 0.768 & 0.777 \\
High ferritin & 0.870 & 0.865 & 0.864 & 0.871 & 0.834 & 0.833 & 0.811 & 0.785 & 0.833 & 0.830 & 0.831 & 0.819 \\
High creatinine kinase & 0.905 & 0.901 & 0.904 & 0.912 & 0.862 & 0.853 & 0.817 & 0.755 & 0.809 & 0.814 & 0.806 & 0.804 \\
High C-reactive protein & 0.849 & 0.853 & 0.851 & 0.855 & 0.814 & 0.807 & 0.785 & 0.742 & 0.793 & 0.786 & 0.799 & 0.758 \\
High erythrocyte sedimentation rate & 0.932 & 0.934 & 0.933 & 0.936 & 0.884 & 0.885 & 0.856 & 0.823 & -\/- & -\/- & -\/- & -\/- \\
Low PaO2 & 0.954 & 0.956 & 0.954 & 0.952 & 0.943 & 0.946 & 0.920 & 0.901 & -\/- & -\/- & -\/- & -\/- \\
Low SpO2 & 0.815 & 0.810 & 0.804 & 0.805 & 0.766 & 0.764 & 0.739 & 0.684 & -\/- & -\/- & -\/- & -\/- \\
\textbf{Medications} & & & & & & & & & & & & \\
Any antibacterial & 0.864 & 0.863 & 0.862 & 0.858 & 0.831 & 0.817 & 0.782 & 0.726 & -\/- & -\/- & -\/- & -\/- \\
Any antifungal & 0.942 & 0.945 & 0.945 & 0.934 & 0.925 & 0.919 & 0.904 & 0.884 & -\/- & -\/- & -\/- & -\/- \\
Any chemotherapy & 0.974 & 0.968 & 0.972 & 0.966 & 0.952 & 0.948 & 0.938 & 0.915 & -\/- & -\/- & -\/- & -\/- \\
Any antiepileptics & 0.870 & 0.862 & 0.862 & 0.860 & 0.832 & 0.821 & 0.791 & 0.733 & -\/- & -\/- & -\/- & -\/- \\
Any glucocorticoid & 0.842 & 0.841 & 0.839 & 0.836 & 0.806 & 0.788 & 0.753 & 0.719 & -\/- & -\/- & -\/- & -\/- \\
Dexamethasone & 0.842 & 0.839 & 0.844 & 0.837 & 0.802 & 0.797 & 0.756 & 0.725 & -\/- & -\/- & -\/- & -\/- \\
Any opioid & 0.853 & 0.845 & 0.845 & 0.840 & 0.814 & 0.807 & 0.784 & 0.727 & -\/- & -\/- & -\/- & -\/- \\
Morphine & 0.836 & 0.830 & 0.826 & 0.823 & 0.799 & 0.786 & 0.757 & 0.708 & -\/- & -\/- & -\/- & -\/- \\
Fentanyl & 0.843 & 0.837 & 0.837 & 0.839 & 0.807 & 0.796 & 0.778 & 0.733 & -\/- & -\/- & -\/- & -\/- \\
Any inotrope & 0.926 & 0.933 & 0.937 & 0.922 & 0.880 & 0.896 & 0.839 & 0.810 & -\/- & -\/- & -\/- & -\/- \\
\textbf{Clinical outcomes} & & & & & & & & & & & & \\
Long length of stay ($\geq$ 7 days) & 0.834 & 0.828 & 0.827 & 0.830 & 0.795 & 0.778 & 0.762 & 0.724 & 0.811 & 0.814 & 0.807 & 0.796 \\
Readmission within 30 days & 0.793 & 0.793 & 0.794 & 0.789 & 0.775 & 0.777 & 0.768 & 0.744 & 0.833 & 0.826 & 0.832 & 0.814 \\
Mortality & 0.941 & 0.935 & 0.940 & 0.931 & 0.905 & 0.878 & 0.903 & 0.828 & 0.895 & 0.892 & 0.900 & 0.884 \\
\end{longtable}\endgroup

*MIMIC cross-site columns include 36 labels.

Abbreviations: AUROC, area under the receiver operating characteristic curve; MIMIC, Medical Information Mart for Intensive Care; SickKids, The Hospital for Sick Children; SK, SickKids.

\end{landscape}
\clearpage
\subsection*{Supplementary Figure S6. Sample efficiency curves across in-domain, cross-vocabulary, and cross-site settings for PORTER and its text encoder ablations}

\includegraphics[width=\linewidth,height=0.85\textheight,keepaspectratio]{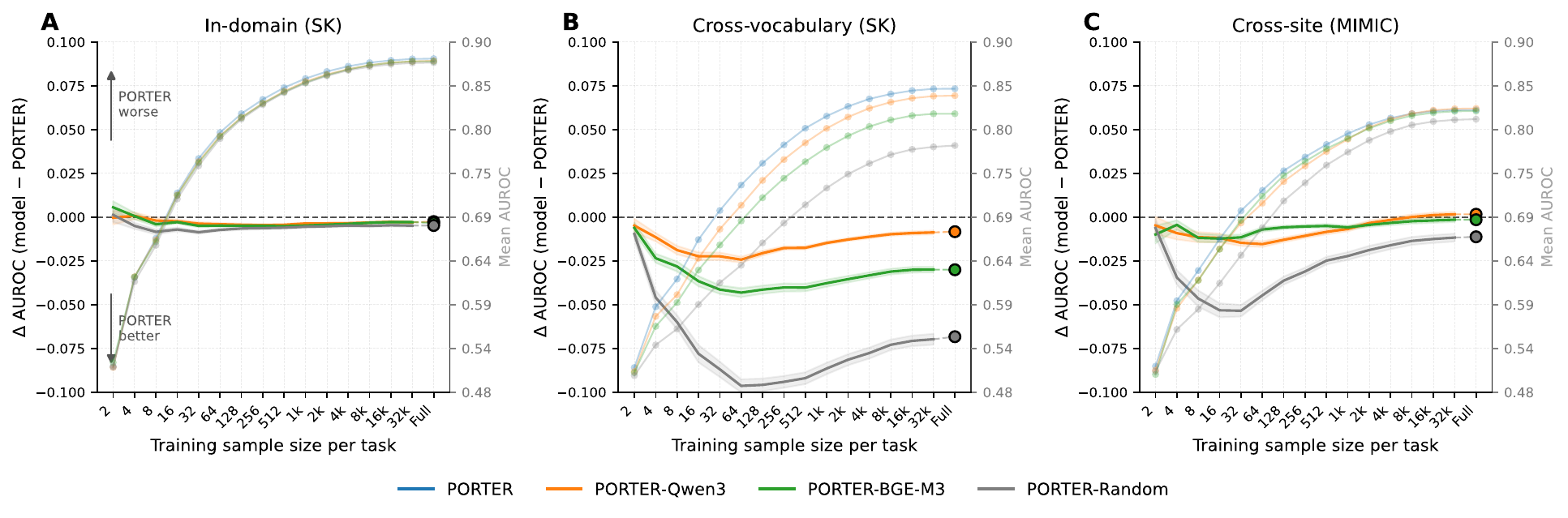}

Curves summarize linear probe performance as labeled training examples per task increase. PORTER is compared against three ablations that hold the FiLM numeric pathway fixed while varying the text encoder: PORTER-Qwen3 (Qwen3-Embedding-8B), PORTER-BGE-M3 (BGE-M3), and PORTER-Random (randomly initialized encoder).

Abbreviations: FiLM, feature-wise linear modulation; FM, foundation model; MIMIC, Medical Information Mart for Intensive Care; SEDAR, SickKids Enterprise-wide Data in Azure Repository; SickKids, The Hospital for Sick Children; SK, SickKids.

\clearpage
\subsection*{Supplementary Figure S7. Patient representation geometry under cross-vocabulary transfer for PORTER and its text encoder ablations}

\includegraphics[width=\linewidth,height=0.85\textheight,keepaspectratio]{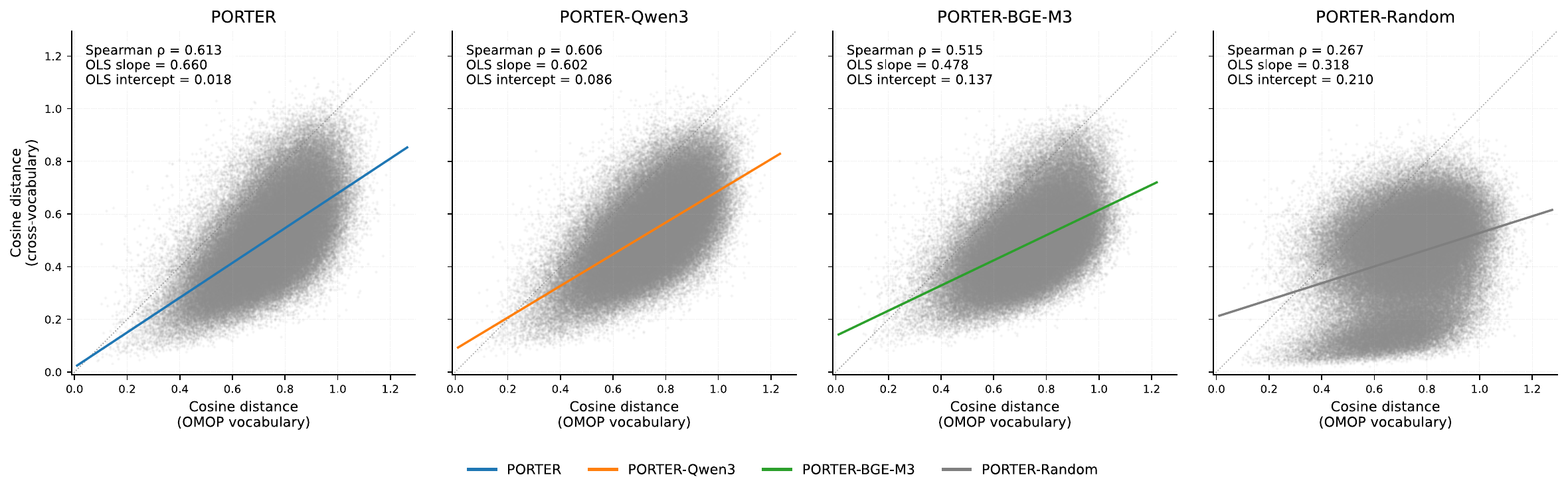}

For each encoder, patient representations were computed for the same 9,585 unique test patients using OMOP-derived event descriptions and SEDAR-derived event descriptions. Each point plots the cosine distance between a pair of patients under the OMOP-derived representation against the corresponding distance under the SEDAR-derived representation. The scatter plots show a random subsample (200,000 pairs) for visualization. Spearman $\rho$ and the OLS fit summarize how well the encoder preserves patient-level geometry. Higher $\rho$, slope closer to 1, and intercept closer to 0 indicate stronger preservation of patient-level geometry under vocabulary transfer.

Abbreviations: OLS, ordinary least squares; OMOP, Observational Medical Outcomes Partnership; SEDAR, SickKids Enterprise-wide Data in Azure Repository.

\textbf{\hfill\break
}

\begin{landscape}
\subsection*{Supplementary Table S10. Per-task AUROC for PORTER and its numeric pathway ablations across in-domain, cross-vocabulary, and cross-site settings}

\begingroup\scriptsize\setlength{\tabcolsep}{4pt}\begin{longtable}[]{@{}
  >{\raggedright\arraybackslash}p{(\columnwidth - 18\tabcolsep) * \real{0.1610}}
  >{\centering\arraybackslash}p{(\columnwidth - 18\tabcolsep) * \real{0.0890}}
  >{\centering\arraybackslash}p{(\columnwidth - 18\tabcolsep) * \real{0.0805}}
  >{\centering\arraybackslash}p{(\columnwidth - 18\tabcolsep) * \real{0.0890}}
  >{\centering\arraybackslash}p{(\columnwidth - 18\tabcolsep) * \real{0.1186}}
  >{\centering\arraybackslash}p{(\columnwidth - 18\tabcolsep) * \real{0.0805}}
  >{\centering\arraybackslash}p{(\columnwidth - 18\tabcolsep) * \real{0.0890}}
  >{\centering\arraybackslash}p{(\columnwidth - 18\tabcolsep) * \real{0.1144}}
  >{\centering\arraybackslash}p{(\columnwidth - 18\tabcolsep) * \real{0.0805}}
  >{\centering\arraybackslash}p{(\columnwidth - 18\tabcolsep) * \real{0.0890}}@{}}
\toprule\noalign{}
\begin{minipage}[b]{\linewidth}\raggedright
\textbf{Task}
\end{minipage} & \multicolumn{3}{c}{\textbf{In-domain (SK)}} & \multicolumn{3}{c}{\textbf{Cross-vocabulary (SK)}} & \multicolumn{3}{c}{\textbf{Cross-site (MIMIC)*}} \\
 & \textbf{PORTER} & \textbf{PORTER-NoNum} & \textbf{PORTER-NumText} & \textbf{PORTER} & \textbf{PORTER-NoNum} & \textbf{PORTER-NumText} & \textbf{PORTER} & \textbf{PORTER-NoNum} & \textbf{PORTER-NumText} \\
\midrule\noalign{}
\endhead
\bottomrule\noalign{}
\endlastfoot
\textbf{Transfusions} & & & & & & & & & \\
Platelet transfusion & 0.967 & 0.963 & 0.968 & 0.944 & 0.946 & 0.942 & -\/- & -\/- & -\/- \\
Red cell transfusion & 0.929 & 0.912 & 0.919 & 0.891 & 0.880 & 0.881 & -\/- & -\/- & -\/- \\
\textbf{Procedure} & & & & & & & & & \\
Invasive intubation & 0.924 & 0.935 & 0.919 & 0.894 & 0.885 & 0.885 & -\/- & -\/- & -\/- \\
Gastrostomy tube & 0.885 & 0.872 & 0.862 & 0.855 & 0.795 & 0.854 & -\/- & -\/- & -\/- \\
Echocardiogram & 0.894 & 0.887 & 0.893 & 0.869 & 0.864 & 0.856 & -\/- & -\/- & -\/- \\
Pulmonary function test & 0.966 & 0.965 & 0.965 & 0.941 & 0.929 & 0.928 & -\/- & -\/- & -\/- \\
Lumbar puncture & 0.947 & 0.934 & 0.950 & 0.916 & 0.914 & 0.909 & -\/- & -\/- & -\/- \\
Surgery & 0.891 & 0.888 & 0.885 & 0.850 & 0.859 & 0.844 & -\/- & -\/- & -\/- \\
Interventional radiology & 0.864 & 0.858 & 0.858 & 0.815 & 0.812 & 0.815 & -\/- & -\/- & -\/- \\
\textbf{Imaging} & & & & & & & & & \\
Plain radiography chest & 0.820 & 0.805 & 0.818 & 0.787 & 0.759 & 0.767 & -\/- & -\/- & -\/- \\
Ultrasound abdomen & 0.841 & 0.836 & 0.843 & 0.810 & 0.799 & 0.802 & -\/- & -\/- & -\/- \\
Computerized tomography chest & 0.868 & 0.860 & 0.864 & 0.818 & 0.815 & 0.818 & -\/- & -\/- & -\/- \\
Computerized tomography abdomen & 0.894 & 0.878 & 0.885 & 0.824 & 0.839 & 0.832 & -\/- & -\/- & -\/- \\
Computerized tomography head & 0.910 & 0.909 & 0.914 & 0.869 & 0.885 & 0.870 & -\/- & -\/- & -\/- \\
MRI head & 0.908 & 0.908 & 0.906 & 0.875 & 0.877 & 0.867 & -\/- & -\/- & -\/- \\
MRI whole body & 0.948 & 0.942 & 0.933 & 0.903 & 0.846 & 0.860 & -\/- & -\/- & -\/- \\
PET & 0.897 & 0.894 & 0.909 & 0.823 & 0.912 & 0.803 & -\/- & -\/- & -\/- \\
\textbf{Laboratory abnormality} & & & & & & & & & \\
High white blood count & 0.834 & 0.794 & 0.824 & 0.781 & 0.740 & 0.773 & 0.760 & 0.747 & 0.745 \\
Low white blood count & 0.920 & 0.888 & 0.916 & 0.883 & 0.864 & 0.878 & 0.869 & 0.844 & 0.844 \\
High absolute neutrophil count & 0.839 & 0.809 & 0.830 & 0.788 & 0.761 & 0.788 & 0.795 & 0.788 & 0.782 \\
Low absolute neutrophil count & 0.934 & 0.915 & 0.935 & 0.906 & 0.898 & 0.908 & 0.853 & 0.844 & 0.822 \\
High bands & 0.872 & 0.864 & 0.868 & 0.845 & 0.830 & 0.834 & -\/- & -\/- & -\/- \\
High lymphocyte & 0.852 & 0.822 & 0.846 & 0.816 & 0.793 & 0.803 & 0.713 & 0.694 & 0.682 \\
Low lymphocyte & 0.894 & 0.884 & 0.892 & 0.871 & 0.859 & 0.864 & 0.816 & 0.805 & 0.791 \\
High hemoglobin & 0.861 & 0.837 & 0.850 & 0.818 & 0.802 & 0.791 & -\/- & -\/- & -\/- \\
Low hemoglobin & 0.887 & 0.867 & 0.880 & 0.855 & 0.830 & 0.841 & 0.859 & 0.847 & 0.851 \\
High mean corpuscular volume & 0.892 & 0.869 & 0.891 & 0.860 & 0.835 & 0.845 & 0.793 & 0.755 & 0.752 \\
Low mean corpuscular volume & 0.856 & 0.804 & 0.856 & 0.779 & 0.759 & 0.765 & 0.806 & 0.680 & 0.663 \\
High reticulocyte count & 0.898 & 0.889 & 0.889 & 0.855 & 0.843 & 0.848 & -\/- & -\/- & -\/- \\
Low reticulocyte count & 0.893 & 0.864 & 0.877 & 0.851 & 0.826 & 0.834 & -\/- & -\/- & -\/- \\
High platelet & 0.841 & 0.818 & 0.841 & 0.805 & 0.779 & 0.792 & 0.792 & 0.770 & 0.775 \\
Low platelet & 0.877 & 0.868 & 0.886 & 0.846 & 0.833 & 0.840 & 0.813 & 0.798 & 0.800 \\
High immature platelet fraction & 0.892 & 0.884 & 0.893 & 0.855 & 0.852 & 0.854 & -\/- & -\/- & -\/- \\
Low immature platelet fraction & 0.870 & 0.862 & 0.872 & 0.844 & 0.841 & 0.837 & -\/- & -\/- & -\/- \\
High mean platelet volume & 0.902 & 0.894 & 0.902 & 0.880 & 0.870 & 0.867 & -\/- & -\/- & -\/- \\
Low mean platelet volume & 0.792 & 0.768 & 0.796 & 0.762 & 0.733 & 0.755 & -\/- & -\/- & -\/- \\
High fibrinogen & 0.891 & 0.874 & 0.885 & 0.848 & 0.845 & 0.836 & 0.753 & 0.744 & 0.719 \\
Low fibrinogen & 0.920 & 0.912 & 0.908 & 0.886 & 0.879 & 0.860 & 0.866 & 0.844 & 0.852 \\
High partial thromboplastin time & 0.899 & 0.892 & 0.899 & 0.865 & 0.853 & 0.862 & 0.867 & 0.852 & 0.854 \\
High international normalized ratio & 0.869 & 0.860 & 0.863 & 0.834 & 0.820 & 0.826 & 0.884 & 0.876 & 0.877 \\
High sodium & 0.851 & 0.836 & 0.846 & 0.813 & 0.799 & 0.805 & 0.818 & 0.795 & 0.798 \\
Low sodium & 0.854 & 0.838 & 0.851 & 0.832 & 0.805 & 0.824 & 0.825 & 0.798 & 0.800 \\
High potassium & 0.836 & 0.826 & 0.831 & 0.807 & 0.792 & 0.799 & 0.834 & 0.817 & 0.821 \\
Low potassium & 0.868 & 0.862 & 0.869 & 0.838 & 0.831 & 0.832 & 0.819 & 0.807 & 0.800 \\
High glucose & 0.832 & 0.824 & 0.831 & 0.796 & 0.788 & 0.789 & 0.927 & 0.924 & 0.925 \\
Low glucose & 0.882 & 0.878 & 0.877 & 0.855 & 0.839 & 0.851 & 0.783 & 0.777 & 0.760 \\
High creatinine & 0.914 & 0.897 & 0.901 & 0.883 & 0.861 & 0.875 & 0.862 & 0.839 & 0.833 \\
High urea & 0.926 & 0.916 & 0.926 & 0.896 & 0.892 & 0.889 & 0.877 & 0.860 & 0.860 \\
Low albumin & 0.878 & 0.870 & 0.874 & 0.849 & 0.843 & 0.841 & 0.849 & 0.839 & 0.836 \\
High alanine transaminase & 0.885 & 0.870 & 0.883 & 0.853 & 0.835 & 0.839 & 0.800 & 0.783 & 0.780 \\
High aspartate aminotransferase & 0.894 & 0.882 & 0.895 & 0.862 & 0.843 & 0.854 & 0.800 & 0.781 & 0.773 \\
High lactate dehydrogenase & 0.913 & 0.902 & 0.898 & 0.875 & 0.868 & 0.860 & 0.827 & 0.811 & 0.806 \\
High bilirubin & 0.893 & 0.887 & 0.884 & 0.862 & 0.850 & 0.855 & 0.834 & 0.800 & 0.814 \\
High cholesterol & 0.906 & 0.883 & 0.898 & 0.864 & 0.809 & 0.855 & 0.785 & 0.756 & 0.804 \\
High triglyceride & 0.876 & 0.865 & 0.858 & 0.817 & 0.801 & 0.772 & 0.773 & 0.758 & 0.753 \\
High ferritin & 0.870 & 0.852 & 0.856 & 0.834 & 0.831 & 0.825 & 0.833 & 0.808 & 0.802 \\
High creatinine kinase & 0.905 & 0.899 & 0.903 & 0.862 & 0.838 & 0.845 & 0.809 & 0.783 & 0.797 \\
High C-reactive protein & 0.849 & 0.838 & 0.844 & 0.814 & 0.793 & 0.802 & 0.793 & 0.790 & 0.781 \\
High erythrocyte sedimentation rate & 0.932 & 0.914 & 0.929 & 0.884 & 0.880 & 0.890 & -\/- & -\/- & -\/- \\
Low PaO2 & 0.954 & 0.960 & 0.956 & 0.943 & 0.946 & 0.948 & -\/- & -\/- & -\/- \\
Low SpO2 & 0.815 & 0.804 & 0.813 & 0.766 & 0.763 & 0.765 & -\/- & -\/- & -\/- \\
\textbf{Medications} & & & & & & & & & \\
Any antibacterial & 0.864 & 0.858 & 0.861 & 0.831 & 0.823 & 0.816 & -\/- & -\/- & -\/- \\
Any antifungal & 0.942 & 0.945 & 0.943 & 0.925 & 0.923 & 0.917 & -\/- & -\/- & -\/- \\
Any chemotherapy & 0.974 & 0.974 & 0.973 & 0.952 & 0.951 & 0.953 & -\/- & -\/- & -\/- \\
Any antiepileptics & 0.870 & 0.863 & 0.864 & 0.832 & 0.816 & 0.813 & -\/- & -\/- & -\/- \\
Any glucocorticoid & 0.842 & 0.836 & 0.840 & 0.806 & 0.800 & 0.797 & -\/- & -\/- & -\/- \\
Dexamethasone & 0.842 & 0.836 & 0.847 & 0.802 & 0.802 & 0.794 & -\/- & -\/- & -\/- \\
Any opioid & 0.853 & 0.845 & 0.849 & 0.814 & 0.814 & 0.808 & -\/- & -\/- & -\/- \\
Morphine & 0.836 & 0.834 & 0.836 & 0.799 & 0.791 & 0.785 & -\/- & -\/- & -\/- \\
Fentanyl & 0.843 & 0.837 & 0.841 & 0.807 & 0.808 & 0.803 & -\/- & -\/- & -\/- \\
Any inotrope & 0.926 & 0.931 & 0.924 & 0.880 & 0.896 & 0.890 & -\/- & -\/- & -\/- \\
\textbf{Clinical outcomes} & & & & & & & & & \\
Long length of stay ($\geq$ 7 days) & 0.834 & 0.823 & 0.831 & 0.795 & 0.777 & 0.782 & 0.811 & 0.805 & 0.798 \\
Readmission within 30 days & 0.793 & 0.791 & 0.790 & 0.775 & 0.778 & 0.775 & 0.833 & 0.836 & 0.825 \\
Mortality & 0.941 & 0.929 & 0.932 & 0.905 & 0.889 & 0.873 & 0.895 & 0.877 & 0.868 \\
\end{longtable}\endgroup

*MIMIC cross-site columns include 36 labels.

Abbreviations: AUROC, area under the receiver operating characteristic curve; MIMIC, Medical Information Mart for Intensive Care; NoNum, no numeric pathway; NumText, numeric-as-text; SickKids, The Hospital for Sick Children; SK, SickKids.

\textbf{\hfill\break
}

\end{landscape}
\clearpage
\subsection*{Supplementary Figure S8. Sample efficiency curves across in-domain, cross-vocabulary, and cross-site settings for PORTER and its numeric pathway ablations}

\includegraphics[width=\linewidth,height=0.85\textheight,keepaspectratio]{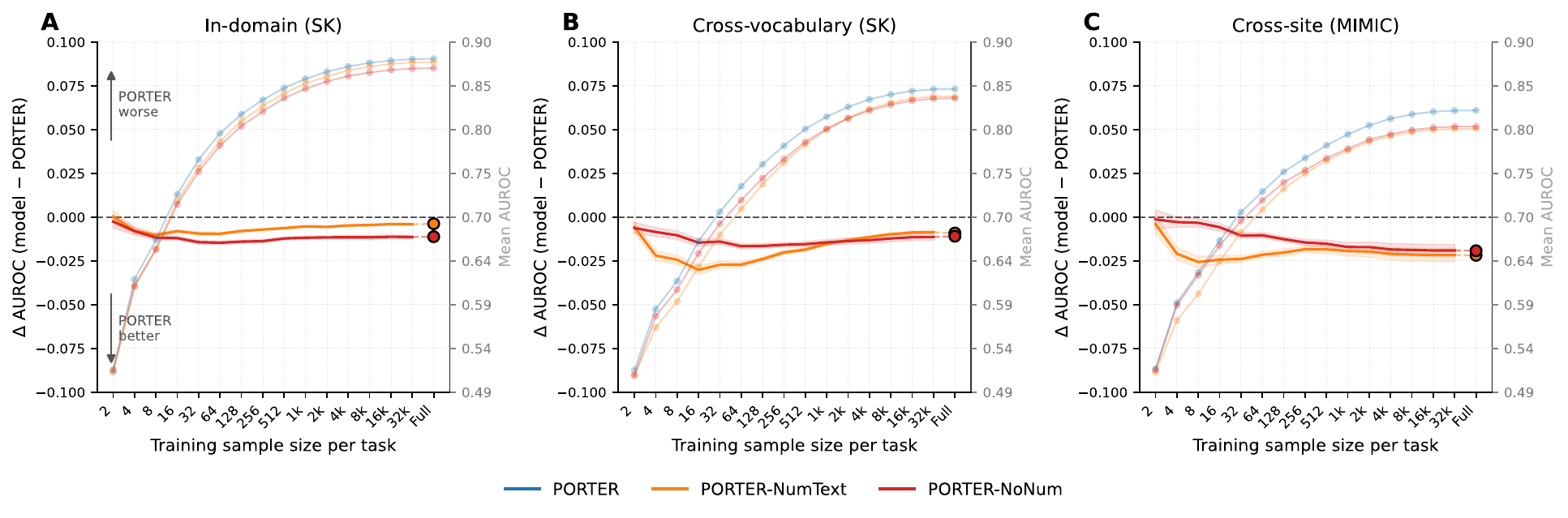}

Curves summarize linear probe performance as labeled training examples per task increase. PORTER is compared against two numeric pathway ablations while holding the text encoder fixed: PORTER-NumText (numeric value and reference range added to the event text description instead of using FiLM) and PORTER-NoNum (no numeric representation).

Abbreviations: FiLM, feature-wise linear modulation; FM, foundation model; MIMIC, Medical Information Mart for Intensive Care; SEDAR, SickKids Enterprise-wide Data in Azure Repository; SickKids, The Hospital for Sick Children; SK, SickKids.

\textbf{\hfill\break
}

\clearpage
\subsection*{Supplementary Figure S9. Synonym invariance and numeric sensitivity of event representations passed to the transformer for PORTER and its numeric pathway ablations}

\includegraphics[width=\linewidth,height=0.85\textheight,keepaspectratio]{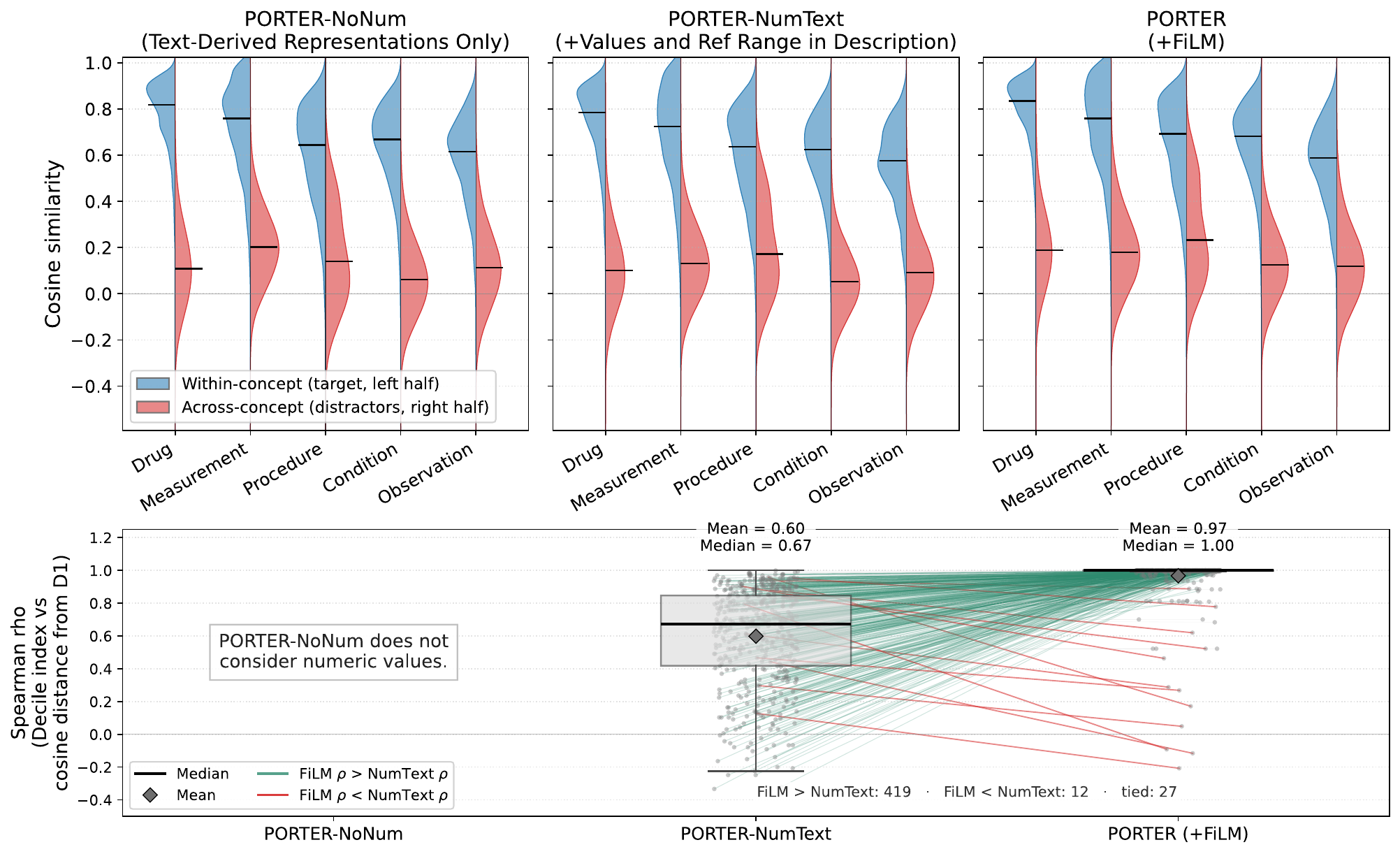}

Top row: cosine similarity between SEDAR-derived descriptions and OMOP-derived event descriptions, shown as half-violins per domain (within-concept on the left, across-concept distractors on the right). Bottom row: Spearman $\rho$ for each measurement concept between decile index and cosine distance from the first decile, shown as box plots with jittered per-concept points. Lines connect the same measurement concept across the PORTER-NumText and PORTER columns. Line color indicates whether FiLM improves or worsens numeric sensitivity relative to NumText, as shown in the legend. PORTER-NoNum does not consider numeric values, so embeddings across deciles are identical for each concept and $\rho$ is undefined.

Abbreviations: FiLM, feature-wise linear modulation; NoNum, no numeric pathway; NumText, numeric values rendered as text; OMOP, Observational Medical Outcomes Partnership; SEDAR, SickKids Enterprise-wide Data in Azure Repository.

\textbf{\hfill\break
}

\end{document}